\documentclass[11pt]{article}

\usepackage[final]{acl}

\usepackage{times}
\usepackage{latexsym}

\usepackage[T1]{fontenc}
\usepackage[utf8]{inputenc}

\usepackage{microtype}

\usepackage{inconsolata}
\usepackage{xspace}

\usepackage{graphicx}

\usepackage{algorithmicx}
\usepackage{booktabs}
\usepackage{times,latexsym}
\usepackage{url}
\usepackage[T1]{fontenc}
\usepackage{enumitem}
\usepackage{amssymb}
\usepackage{tikz}
\usetikzlibrary{shapes}
\usepackage{amsmath} 
\usepackage{subfigure}
\usepackage{makecell}
\usepackage{algorithm}
\usepackage{algpseudocode}
\usepackage{tcolorbox}
\usepackage{colortbl}
\usepackage{multirow}
\usepackage{amssymb}
\usepackage[thicklines]{cancel}
\makeatletter
\newcommand{\rmnum}[1]{\romannumeral #1}
\newcommand{\Rmnum}[1]{\expandafter\@slowromancap\romannumeral #1@}
\makeatother
\usepackage{hyperref}
\usepackage{marvosym}
\usepackage{xcolor}

\newcommand{\think}[1]{\textcolor{blue}{\texttt{<think>}} #1 \textcolor{blue}{\texttt{</think>}}}
\newcommand{\search}[1]{\textcolor{cyan}{\texttt{<search>}} #1 \textcolor{cyan}{\texttt{</search>}}}
\newcommand{\info}[1]{\textcolor{brown}{\texttt{<information>}} #1 \textcolor{brown}{\texttt{</information>}}}
\newcommand{\infoEvi}[1]{\textcolor[RGB]{185,154,58}{\texttt{<information w/ \ours{}>}} #1 \textcolor[RGB]{185,154,58}{\texttt{</information w/ \ours{}>}}}
\newcommand{\answer}[1]{\textcolor{purple}{\texttt{<answer>}} #1 \textcolor{purple}{\texttt{</answer>}}}

\newcommand{\ours}{\texttt{EviOmni}\xspace}

\newcommand{\circledoneblackone}{%
  \begin{tikzpicture}[baseline=(char.base)]
    \node[
      shape=circle,
      fill=black,
      draw=black,
      inner sep=0.1pt,
      text=white
    ] (char) {\textbf{1}};
  \end{tikzpicture}%
}

\newcommand{\circledoneblacktwo}{%
  \begin{tikzpicture}[baseline=(char.base)]
    \node[
      shape=circle,
      fill=black,
      draw=black,
      inner sep=0.1pt,
      text=white
    ] (char) {\textbf{2}};
  \end{tikzpicture}%
}

\newcommand{\smallsection}[1]{{\vspace{0.07in} \noindent \bf {#1.\hspace{5pt}}}}

\title{Learning to Extract Rational Evidence via Reinforcement Learning \\ for Retrieval-Augmented Generation}

\author{Xinping Zhao$^{1,2}$,  Shouzheng Huang$^{1}$, Yan Zhong$^{3}$, Xinshuo Hu,  \\ \textbf{Meishan Zhang$^{1}$\textsuperscript{\Letter}\thanks{\textsuperscript{\Letter}Corresponding author.}, Baotian Hu$^{1,2}$,  Min Zhang$^{1,2}$} \\ 
        $^{1}$Harbin Institute of Technology (Shenzhen), $^{2}$Shenzhen Loop Area Institute, $^{3}$Peking University \\
        \texttt{\{zhaoxinping, 210110129\}@stu.hit.edu.cn}, \\
        \texttt{zhongyan@stu.pku.edu.cn}, \texttt{yanshek.woo@gmail.com}, \\
          \texttt{mason.zms@gmail.com}, \texttt{\{hubaotian, zhangmin2021\}@hit.edu.cn}}
\makeatletter
\def\thanks#1{\protected@xdef\@thanks{\@thanks
        \protect\footnotetext{#1}}}
\makeatother

\begin{document}
\maketitle
\begin{abstract}
  Retrieval-Augmented Generation (RAG) effectively improves the accuracy of Large Language Models (LLMs).
  However, retrieval noises significantly undermine the quality of LLMs' generation, necessitating the development of denoising mechanisms. 
  Previous works extract evidence straightforwardly without deep thinking, which may risk filtering out key clues and struggle with generalization.
  To this end, we propose \ours{}, which learns to extract rational evidence via reasoning first and then extracting.
  %, to avoid omitting any key cues helpful for generation
  Specifically, \ours{} integrates evidence reasoning and evidence extraction into one unified trajectory, followed by knowledge token masking to avoid information leakage, optimized via on-policy reinforcement learning with verifiable rewards in terms of \textit{answer}, \textit{length}, and \textit{format}.
  Extensive experiments on five benchmark datasets show the superiority of \ours{}, which provides compact and high-quality evidence, enhances the accuracy of downstream tasks, and supports both traditional and agentic RAG systems\footnote{Models are available at \url{https://huggingface.co/HIT-TMG/EviOmni-nq_train-1.5B} and \url{https://huggingface.co/HIT-TMG/EviOmni-nq_train-7B}. Code is available at \url{https://github.com/HITsz-TMG/EviOmni}.}.
\end{abstract}

\section{Introduction}
\label{sec:intro}
Retrieval-Augmented Generation (RAG) prevails in Large Language Models (LLMs). 
It has demonstrated strong effectiveness across a wide array of knowledge-intensive tasks \cite{DBLP:conf/nips/LewisPPPKGKLYR020,DBLP:conf/emnlp/WuZHMS022}, such as open-domain question answering (QA) \cite{DBLP:conf/naacl/ShiMYS0LZY24,DBLP:conf/acl/TrivediBKS23}, fact-checking \cite{DBLP:conf/emnlp/DuZWCBQC023,zhao2024medico}, and dialog generation \cite{DBLP:journals/jmlr/IzacardLLHPSDJRG23,DBLP:journals/corr/abs-2201-08239}, to produce faithful, reliable, and accurate outputs. 
Ideally, LLMs should be grounded in purely relevant content to generate accurate output and facilitate inference speed.
However, due to imperfect retrieval systems and noisy retrieval corpus \cite{DBLP:journals/corr/abs-2311-08377,DBLP:journals/corr/abs-2406-13629,DBLP:conf/emnlp/ZhaoLZHCHZ24}, retrieval passages often contain many irrelevant or noisy snippets, distracting LLMs' attention and degrading generation quality.
As such, it is necessary to extract evidence and filter out noise for RAG to achieve superior performance.
\begin{figure}[t]
    \centering
    \includegraphics[width=1.\linewidth]{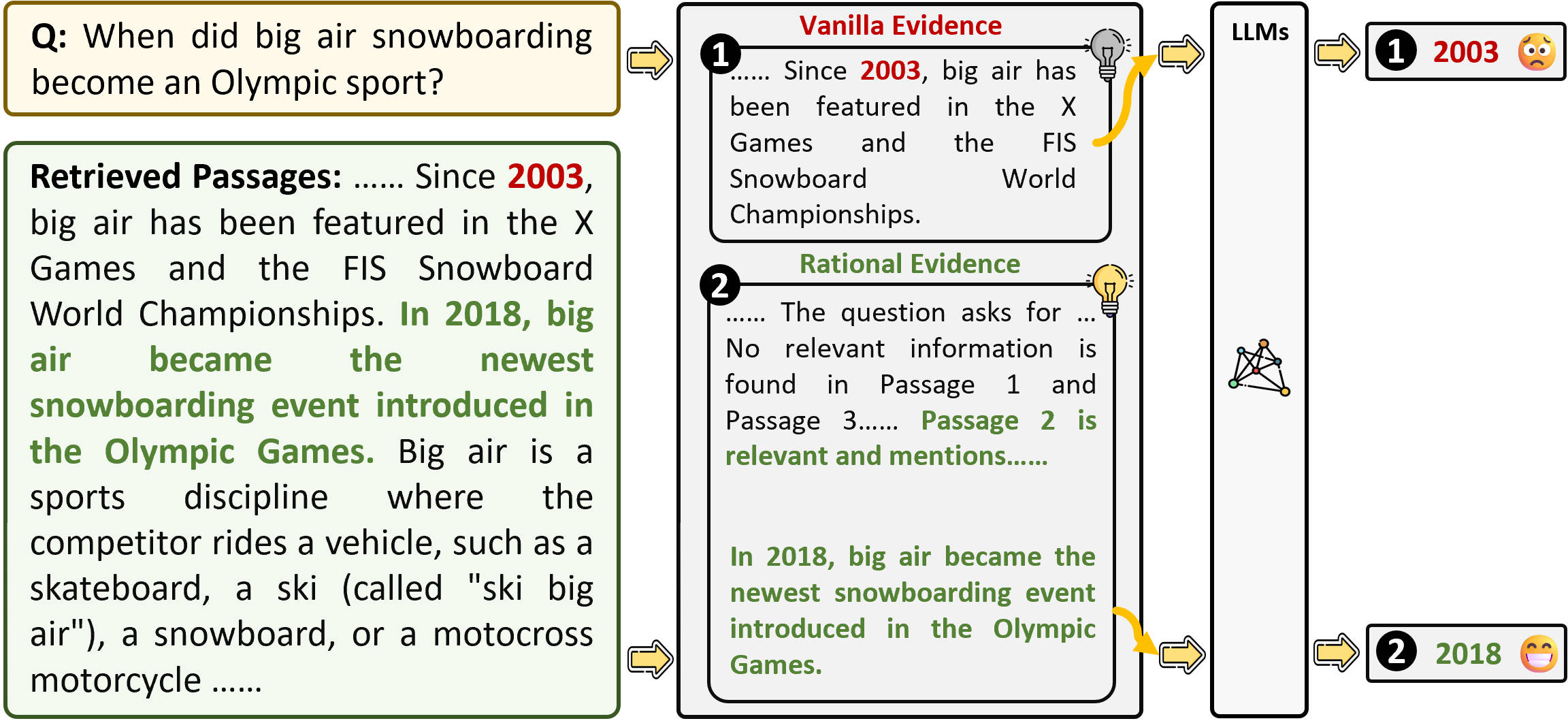}
    \caption{Motivating example, where key clues are marked in \textbf{\textcolor[RGB]{0,100,0}{green}}: \protect\circledoneblackone{} The key clue is omitted, leading to an incorrect answer; \protect\circledoneblacktwo{} Guided by evidence reasoning, the key clue is extracted, leading to a correct answer.} 
    \label{fig:motivation}
\end{figure}
Recently, several studies have attempted to address this issue. 
They can be broadly divided into two categories: \textbf{(1)} Reranking more relevant passages to the top of the retrieval list~\cite{DBLP:journals/corr/abs-2407-03627,DBLP:journals/corr/abs-2501-04695,DBLP:conf/acl/MaoHLSGHC21}; and \textbf{(2)} Summarizing retrieval contents into a coherent and relevant narrative~\cite{DBLP:journals/corr/abs-2311-08377,DBLP:conf/emnlp/ZhaoLZHCHZ24,DBLP:conf/acl/0025FDGY0CCC024,DBLP:conf/iclr/XuSC24}. 
The former heavily relies on the performance of the reranker itself and may disrupt context after reranking. 
It usually performs worse than the latter due to a lack of contextual understanding.
The latter aims to train LLMs as filtering models via supervised fine-tuning (SFT)~\cite{DBLP:journals/corr/abs-2311-08377} or preference optimization \cite{DBLP:conf/emnlp/ZhaoLZHCHZ24}. 
However, these methods typically rely on heuristically constructed training data, \textit{e.g.,} String Inclusion that measures whether the golden answer is included in the passage, and Lexical Overlap that calculates the unigram overlap between the answer and passage. 
Despite their effectiveness, existing methods rerank or summarize retrieval contents straightforwardly without deep thinking, which may risk filtering out key clues due to insufficient contextual understanding of retrieved contents.
Figure \ref{fig:motivation} presents a comparison between vanilla evidence and rational evidence for QA:
\circledoneblackone{} The vanilla evidence omits the key clue and leads to an incorrect answer;
and \circledoneblacktwo{} In contrast, evidence reasoning guides the subsequent evidence extraction, \textit{i.e.,} ``\textit{Passage 2 is relevant and mentions......}'', so the rational evidence preserves the key clue, leading to a correct answer.
This finding motivates us to develop rational evidence extraction for RAG, which first performs deep reasoning and then consciously extracts evidence.
In this work, we propose \ours{}, a framework that learns to extract rational evidence for RAG via reinforcement learning with verifiable rewards.
Specifically, \ours{} integrates evidence reasoning (enclosed within \textbf{\textcolor[RGB]{68,114,196}{<reason>}} and \textbf{\textcolor[RGB]{68,114,196}{</reason>}}) and extraction (enclosed within \textbf{\textcolor[RGB]{0,100,0}{<extract>}} and \textbf{\textcolor[RGB]{0,100,0}{</extract>}}) into one response enabling end-to-end training.
The model generates multiple responses to collect informative trajectories containing both positive and negative experiences.
Subsequently, \ours{} applies knowledge token masking on each response to avoid information leakage when generating rationale-based and evidence-based answers (enclosed within \textbf{\textcolor[RGB]{127,100,0}{<answer>}} and \textbf{\textcolor[RGB]{127,100,0}{</answer>}}) to assess their quality, respectively.
Finally, we design three types of rule-based verifiable reward functions, \textit{i.e.,} \textit{answer}, \textit{length}, and \textit{format}, to guide model optimization via Group Relative Policy Optimization (GRPO) \cite{DBLP:journals/corr/abs-2402-03300} towards the desired properties. 
As a result, \ours{} can reason over retrieval contents and consolidate key clues into a concise yet informative narrative, thereby improving LLMs' generation accuracy.
The main contributions of this work can be summarized as:
\begin{itemize}[leftmargin=*]
    \item We unveil and empirically validate the major issue hindering evidence extraction for RAG, \textit{i.e.,} insufficient contextual understanding of retrieval contents, which risks filtering out key clues.
    \item We propose a novel rational evidence extraction learning framework, \ours{}, which first reasons and then extracts, optimized through reinforcement learning with verifiable rewards.
    \item Extensive experiments on five benchmarks show that \ours{} is effective for both traditional and agentic RAG, with superior accuracy, generalization, efficiency, and robustness to retrieval noise.
\end{itemize}
\section{Preliminaries}
\subsection{Problem Statement}
\label{sec:ps}
In RAG~\cite{DBLP:conf/nips/LewisPPPKGKLYR020}, LLMs are given an query $q$ and top-$k$ retrieved passages $P = \{p_1, p_2, ..., p_k\}$, to generate an output $o$ that approximates the golden answer $a$.
In the traditional RAG, retrieved passages $P$ are directly fed into LLMs.
However, these passages may contain irrelevant or noisy contents~\cite{DBLP:journals/corr/abs-2311-08377,DBLP:conf/emnlp/ZhaoLZHCHZ24}, substantially degrading LLMs' generation accuracy and efficiency.
To this end, evidence extraction is introduced to condense $P$ into a concise, query-relevant context $e$, thereby improving generation quality and speed.
The paradigm of ``\textit{RAG with evidence extraction}'' can be formulated as:
\begin{equation}
    e = \mathcal{M}_\mathcal{E}(\cdot|q,P),~~o = \mathcal{M}_\mathcal{G}(\cdot|q,e),
\end{equation}
where $\mathcal{M}_\mathcal{E}(\cdot)$ denotes the evidence extractor; $e$ is the extracted evidence; $\mathcal{M}_\mathcal{G}(\cdot)$ denotes the answer generator. 
Albeit effective, the vanilla paradigm may risk filtering out key clues due to insufficient contextual understanding of retrieved passages.
To address this issue, we propose a novel paradigm of ``\textit{RAG with rational evidence extraction}'':
\begin{equation}
\begin{split}
    e &= \mathcal{M}_\mathcal{E}(\cdot|q,P,r)\cdot\mathcal{M}_\mathcal{E}(r|q,P),  \\
      o &= \mathcal{M}_\mathcal{G}(\cdot|q,e), 
\end{split}
\end{equation}
where $r$ denotes the rationale, which explicitly and thoroughly identifies any clues lying in retrieved passages ($P$) to guide rational evidence extraction ($e$).
We train the model in an on-policy manner, meaning that $\mathcal{M}_\mathcal{E}(\cdot)$ and $\mathcal{M}_\mathcal{G}(\cdot)$ are identical during training.
The overall objective is to learn a policy $\pi_\theta: e \sim \mathcal{M}_\mathcal{E}(\cdot|q,P,r)\cdot\mathcal{M}_\mathcal{E}(r|q,P)$, to extract rational evidence $e$ that maximizes the expected utility $\mathcal{U}(a,o)$ over the data distribution $\mathcal{D}$:
\begin{align}
    \max_{\pi_\theta}~  &\mathbb{E}_{a \sim \mathcal{M}_\mathcal{G}(\cdot|q,e),~ e \sim \pi_\theta(q,P),~ q \sim \mathcal{D}} \Big[\mathcal{U}(a, o)\Big], \notag\\
     &s.t.\  \mathrm{L}_e \ll  \mathrm{L}_P
\end{align}
where $\mathcal{U}(\cdot)$ is a utility function that measures the quality of outputs conditioned on $a$; $\mathrm{L}_e$ and $\mathrm{L}_P$ are the length of evidence and passages, respectively.

\subsection{Empirical Study}
\label{sec:es}
\smallsection{Experimental Setup.} We conduct an empirical study to verify the assumption that evidence extractors can retain as many key clues as possible via reasoning first and then extracting. 
We construct a synthetic instruction dataset by utilizing DeepSeek-R1~\cite{DBLP:journals/corr/abs-2501-12948}, where we sample 1K instances from the training set of Natural Question (NQ)~\cite{DBLP:journals/tacl/KwiatkowskiPRCP19}. 
The output of each instance in the dataset contains three parts: \textbf{(1)} \textbf{\textcolor[RGB]{68,114,196}{<reason>}} evidence reasoning \textbf{\textcolor[RGB]{68,114,196}{</reason>}}; \textbf{(2)} \textbf{\textcolor[RGB]{0,100,0}{<extract>}} evidence extraction \textbf{\textcolor[RGB]{0,100,0}{</extract>}}; and \textbf{(3)} \textbf{\textcolor[RGB]{127,100,0}{<answer>}} final answer \textbf{\textcolor[RGB]{127,100,0}{</answer>}}.
We filter out instances with \textbf{incorrect} final answers, resulting in about 620 instances.
Then, we create two variants of the dataset, where the output of the first one consists of ``\textbf{\textcolor[RGB]{68,114,196}{<reason>}} ... \textbf{\textcolor[RGB]{68,114,196}{</reason>}}\textbf{\textcolor[RGB]{0,100,0}{<extract>}} ... \textbf{\textcolor[RGB]{0,100,0}{</extract>}}''; that of the second one only consists of ``\textbf{\textcolor[RGB]{0,100,0}{<extract>}} ... \textbf{\textcolor[RGB]{0,100,0}{</extract>}}''. 
Finally, we fine-tune Qwen2.5-1.5B-Instruct on these two datasets via Low-Rank Adaptation (LoRA), respectively. The LoRA Rank is set to 16; the number of epochs is set to 3.
We tested on in-distribution NQ and out-of-distribution (OOD) {HotoptQA} \cite{DBLP:conf/emnlp/Yang0ZBCSM18} datasets, excluding instances from the test set, where the retrieved passages did not contain the golden answer.
For a fair comparison, the maximum length of the extracted evidence is set to 100.
\begin{figure}[t]
    \centering  
     \subfigure[Results on NQ.]{
        \includegraphics[width=.48\linewidth]{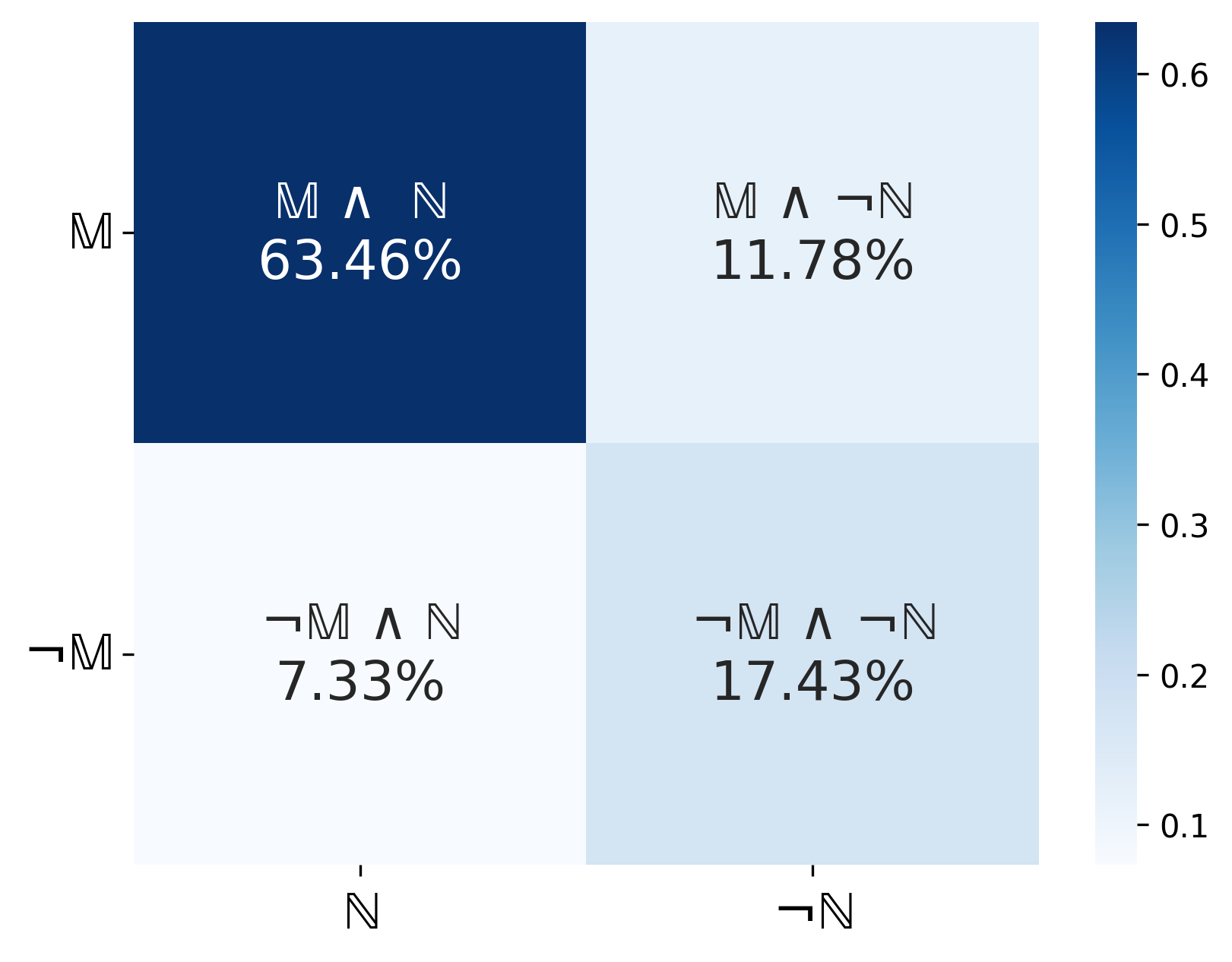}\label{fig:em_study_nq}}
    \subfigure[Results on {HotoptQA}.]{
        \includegraphics[width=.48\linewidth]{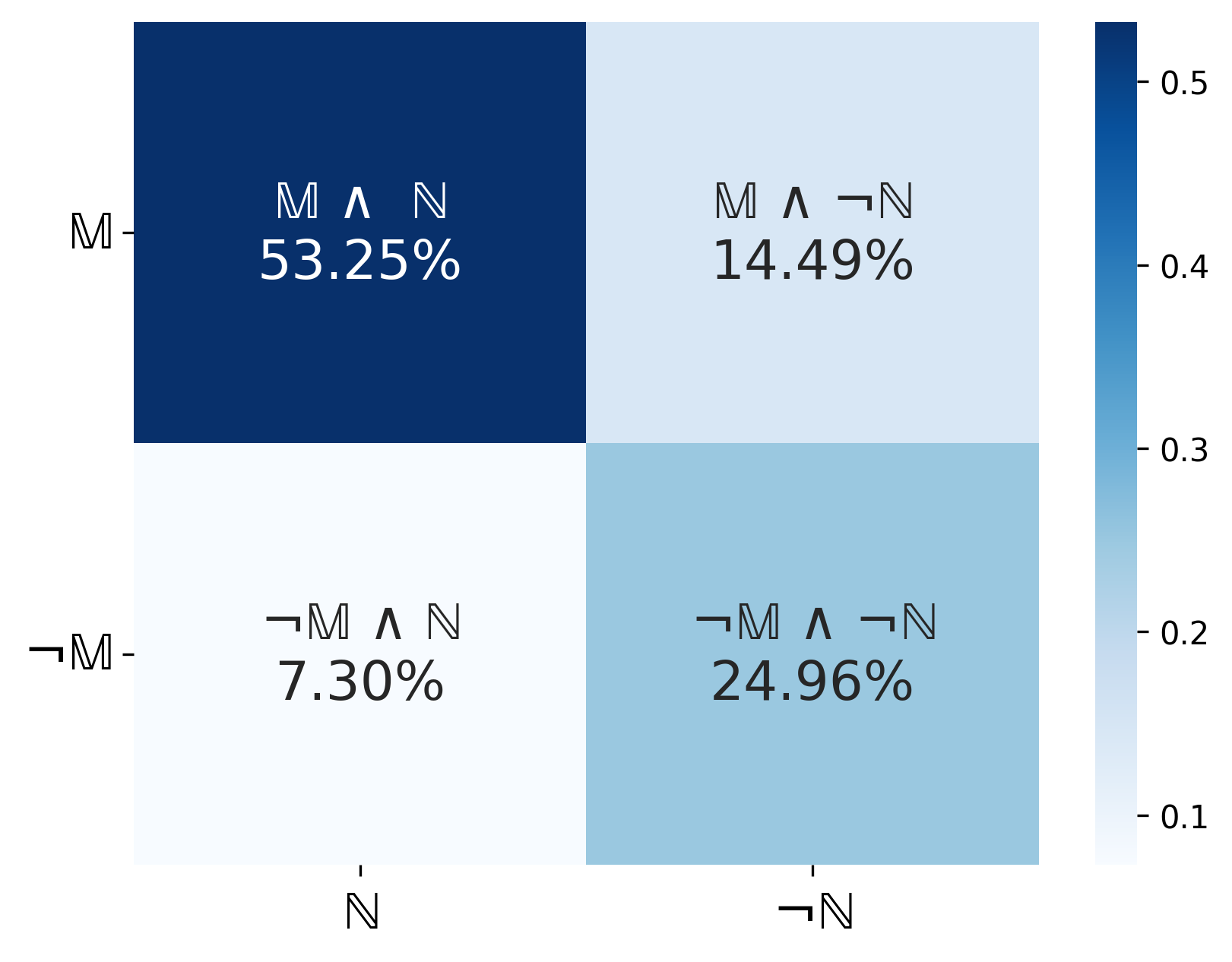}\label{fig:em_study_hqa}}
    \caption{The results \textit{w.r.t.} AR on the in-distribution NQ as well as out-of-distribution {HotoptQA} datasets.}
    \label{fig:em_study}
\end{figure}
\begin{table}[t]
  \centering
  \fontsize{8}{9.5}\selectfont
  \tabcolsep=0.125cm
  \renewcommand\arraystretch{0.99}
  \begin{tabular}{lccc}
    \toprule
    \multicolumn{1}{l}{\textbf{Dataset}} & 
    \multicolumn{1}{c}{\textbf{{Vanilla  Evidence}}} & \multicolumn{1}{c}{\textbf{{Rational Evidence}}} & \multicolumn{1}{c}{\textbf{Rationale}}\\
    \midrule
    NQ    & 70.79\% & 75.24\%  & 77.30\% \\
    HotoptQA   & 60.55\% & 67.74\% & 71.48\% \\
    \bottomrule
  \end{tabular}
  \caption{{AR results of Vanilla Evidence, Rational Evidence, and Rationale on NQ and HotpotQA datasets.}}
  \label{tab:em_study}
\end{table}
\smallsection{Analysis and Discussion} Table \ref{tab:em_study} and Figure \ref{fig:em_study} present the results \textit{w.r.t.} Answer Recall (AR) \cite{DBLP:conf/naacl/ZhaoZSHLLHZ25,DBLP:journals/corr/abs-2406-15319}, measuring the recall of the golden answer string in the evidence. 
The evidence extracted by the model trained on the first and second datasets is denoted as ``rational evidence'' (marked as $\mathrm{\mathbb{M}}$) and ``vanilla evidence'' (marked as $\mathrm{\mathbb{N}}$), respectively.
Here, ``rationale'' refers to the evidence reasoning.
Taking $\mathrm{\mathbb{M}} \land \neg\mathrm{\mathbb{N}}$ as an example, it means the answer string is recalled by the rational evidence but not by the vanilla one.
The results show:
\textbf{(1)} Performance with rational evidence consistently outperforms that with vanilla one, whether on in-distribution or OOD datasets;
\textbf{(2)} Performance with rational evidence is slightly lower than that with rationale, indicating the necessity of reasoning first and better optimization needed to bridge this gap;
and \textbf{(3)} The percentage of $\mathrm{\mathbb{M}} \land \neg\mathrm{\mathbb{N}}$ is considerably higher than that of $\neg\mathrm{\mathbb{M}} \land \mathrm{\mathbb{N}}$,
\textit{e.g.,} 14.49\% vs. 7.30\% on HotpotQA, 
showing the rational evidence's superiority.
\smallsection{Case Study} Detailed case studies on traditional and agentic RAG are provided in Appendices~\ref{app:case} and \ref{app:agen_case}.
Briefly, studies on traditional RAG show that the limitations of vanilla evidence versus rational one lie in the ``\textbf{3I}'' issues: \textbf{(1)} \textbf{I}ncompleteness: lack of key clues; \textbf{(2)} \textbf{I}rrelevance: irrelevant information; and \textbf{(3)} \textbf{I}naccuracy: incorrect information.
In contrast, rational evidence benefits agentic RAG via the ``\textbf{4R}'' properties: \textbf{(1) R}eduction of redundant context; \textbf{(2) R}efusal when evidence is insufficient; \textbf{(3) R}obustness to noise; and \textbf{(4) R}ight timing, which early stops once evidence is enough.
Given the above \textbf{qualitative} and \textbf{quantitative} analyses, it is necessary and valuable to learn to extract rational evidence for improving overall RAG performance.
\begin{figure*}[t]
    \centering
    \includegraphics[width=1.\linewidth]{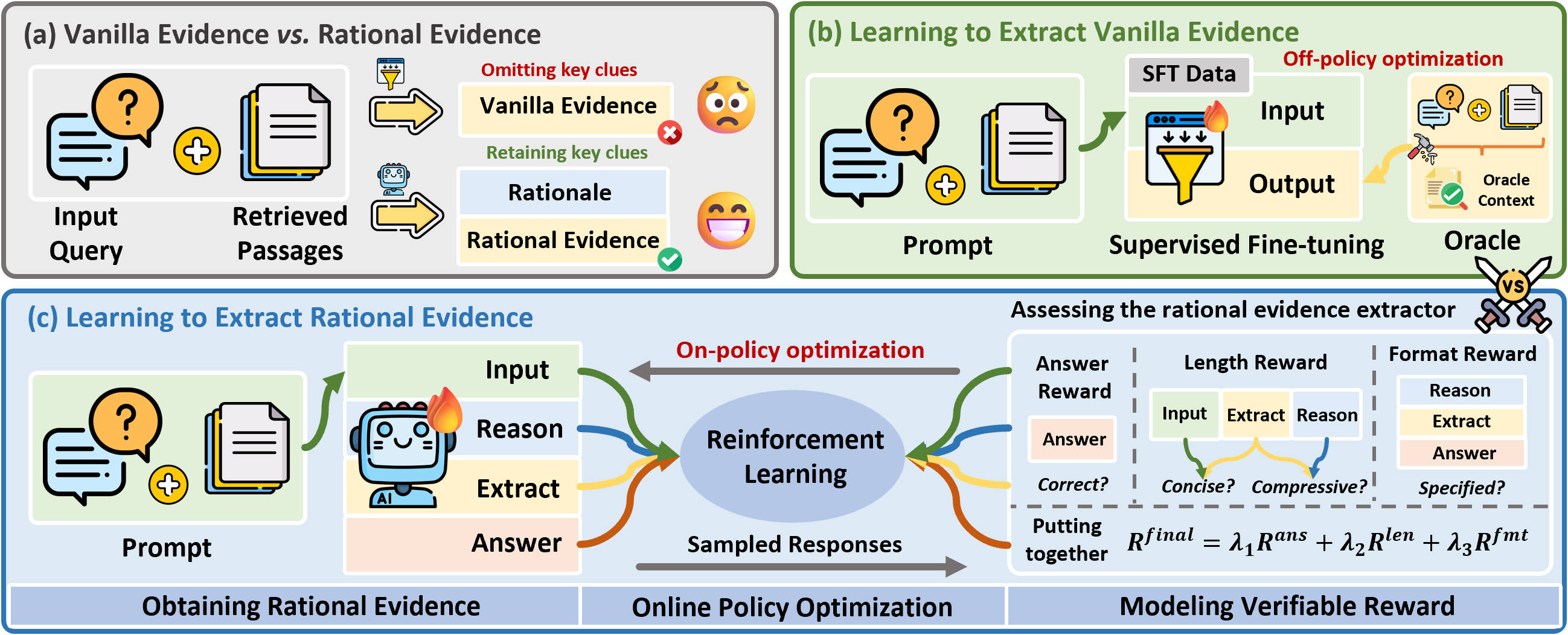}
    \caption{The overall framework of \ours{} and previous vanilla one. \textbf{(a)} Compared to vanilla evidence, rational evidence is more skilled in extracting key clues. \textbf{(b)} Existing works typically learn to extract vanilla evidence via SFT, where the output, \textit{i.e.,} oracle context containing key clues, is usually created by hand-crafted strategies. \textbf{(c)} Our  \ours{} incentivizes rational evidence extraction capability in the extractor $\pi_\theta$ via RL with verifiable rewards.} 
    \label{fig:framework}
\end{figure*}
\section{Methodology}
\label{sec:method}
The overall framework of \ours{} is illustrated in Figure~\ref{fig:framework}
including rational evidence acquisition~(\S\ref{sec:obtain}), modeling verifiable reward~(\S\ref{sec:modeling}), and online policy optimization~(\S\ref{sec:online}).
Algorithms~\ref{alg:trad_rag} and \ref{alg:agent_rag} illustrate how \ours{} enhances the workflows of traditional and agentic RAG, respectively.
\subsection{Obtaining Rational Evidence}
\label{sec:obtain}
\smallsection{Rational Evidence Extraction.} As stated in \S\ref{sec:intro}, vanilla evidence risks filtering out key clues~\cite{DBLP:journals/corr/abs-2311-08377,DBLP:conf/emnlp/ZhaoLZHCHZ24}. An empirical study in \S\ref{sec:es} further verifies our assumption, where rational evidence performs better than vanilla one.
To this end, we aim to optimize the evidence extractor $\mathcal{M}_\mathcal{E}(\cdot)$ to learn to extract rational evidence for RAG, formulated as: $e \sim \mathcal{M}_\mathcal{E}(\cdot|q,P,r)\cdot\mathcal{M}_\mathcal{E}(r|q,P)$. 
Specifically, we feed the query $q$ and its corresponding retrieved passages $P$ into the evidence extractor $\mathcal{M}_\mathcal{E}(\cdot)$ and instruct the evidence extractor to reason first and then extract. 
Evidence reasoning and evidence extraction are enclosed within the special ``reason'' and ``extract'' tags, respectively, \textit{i.e.,} \textbf{\textcolor[RGB]{68,114,196}{<reason>}} ... \textbf{\textcolor[RGB]{68,114,196}{</reason>}} and \textbf{\textcolor[RGB]{0,100,0}{<extract>}} ... \textbf{\textcolor[RGB]{0,100,0}{</extract>}}.
The prompt used for rational evidence generation is provided in Appendix~\ref{app:prompt}.
Given the rationale and the rational evidence, a question arises: \textit{How to assess their quality?}
In \S\ref{sec:es}, we use answer recall, a heuristic metric; similarly, previous works~\cite{DBLP:journals/corr/abs-2311-08377} employ String Inclusion or Lexical Overlap to measure answer recall or unigram overlap, respectively.
However, these metrics do not directly align with the ultimate goal of RAG, namely, generating output as close to the golden answer as possible. 
\smallsection{Knowledge Token Masking} To this end, we propose evaluating the quality of the rationale and rational evidence by assessing the generated outputs that are conditioned on them.
As is well known, causal attention mechanisms in LLMs~\cite{radford2018improving,DBLP:conf/nips/VaswaniSPUJGKP17} aggregate information from preceding contexts, raising the question: \textit{How to generate the output conditioned on them respectively without information leakage?} 
To address this issue, we apply knowledge token masking on each response to disentangle the rationale and rational evidence:
\textbf{(1)} Masking the rational evidence $e$ to assess the quality of the rationale $r$; 
\textbf{(2)} Masking both the retrieved passages $P$ and the rationale $r$ to assess the quality of the rational evidence $e$.
We adopt hard masking on the input rather than soft masking on the attention, because the latter may cause information leakage due to the hidden states already aggregating information from preceding contexts.
After token masking, we first prefill input contexts from scratch and then generate three outputs conditioned on different contexts:
\begin{equation}
\label{equ:out}
\begin{split}
    o_{r} &= \mathcal{M}_{\mathcal{G}}(\cdot|q,P,r,\bcancel{e}), \\
    o_{e} &= \mathcal{M}_{\mathcal{G}}(\cdot|q,\bcancel{P},\bcancel{r},e), \\
    o_{f} &= \mathcal{M}_{\mathcal{G}}(\cdot|q,P,r,e),
\end{split}
\end{equation}
where $o_{r}$, $o_{e}$, and $o_{f}$ represent the outputs conditioned on the rationale, rational evidence, and full context, respectively; $\bcancel{*}$ denotes the hard masking during input. 
Note that we also generate the output $o_{f}$ conditioned on the \underline{f}ull context, since introducing $o_{f}$ into optimization can facilitate convergence. 
\subsection{Modeling Verifiable Reward}
\label{sec:modeling}
The reward functions play a key role in Reinforcement Learning (RL)~\cite{DBLP:conf/nips/Ouyang0JAWMZASR22,DBLP:journals/corr/abs-2412-10400,DBLP:journals/corr/abs-2501-12948}, which can guide the optimization towards desired properties.
To train \ours{} via RL, we design three types of verifiable reward functions \textit{w.r.t.} three primary desired properties for evidence extraction: \textbf{(1)} {The correctness of generated outputs};
\textbf{(2)} {The comprehensiveness of rationale and the conciseness of rational evidence}; and \textbf{(3)} {The format specification of responses}.
For reward modeling, we first collect a set of \textsc{Preao}-tuple $<P, r, e, a, o_{r,e,f}>$. Each \textsc{Preao}-tuple consists of retrieved \underline{P}assgaes $P$, \underline{r}ationale $r$, rational \underline{e}vidence $e$, golden \underline{a}nswer $a$, and generated \underline{o}utput $o_{r,e,f}$.
Then, we design three types of reward functions to assess the rational evidence extractor against the three properties:
\smallsection{Answer Reward} 
It focuses on the correctness of the outputs enclosed within \textbf{\textcolor[RGB]{127,100,0}{<answer>}} and \textbf{\textcolor[RGB]{127,100,0}{</answer>}} tags. 
However, different downstream tasks (\textit{e.g.,} QA, fact verification, and dialog generation) of RAG use different metrics to evaluate correctness, \textit{e.g.,} Exact Match (EM) and $\mathrm{F}_1$ score. 
Given that, a question naturally arises: \textit{How to uniformly measure the answer reward across various RAG downstream tasks?} 
To address this issue, we employ the unigram $\mathrm{F}_1$ to measure the answer rewards in a unified and balanced manner: $R_{*}^{ans} = \mathrm{F}_1(a, o_*), * \in\{r,e,f\}$,
where $R_{*}^{ans} \in [0, 1]$ is the answer reward for the output $o_*$. If $o_*$ is analogous to $a$, then $R_{*}^{ans}$ is close to 1.0; otherwise, 0.0.
\smallsection{Length Reward}
It focuses on two aspects.
The first one is the comprehensiveness of rationale, enclosed within \textbf{\textcolor[RGB]{68,114,196}{<reason>}} and \textbf{\textcolor[RGB]{68,114,196}{</reason>}} tags, where the rationale typically needs to be relatively long to identify any clues lying in retrieval contents. 
The second one is the conciseness of rational evidence, enclosed within \textbf{\textcolor[RGB]{0,100,0}{<extract>}} and \textbf{\textcolor[RGB]{0,100,0}{</extract>}} tags. In contrast, the rational evidence needs to be relatively short to accelerate inference speed.
To quantify ``long'' or ``short'', a reference system is required.
Following prior work~\cite{DBLP:journals/corr/abs-2311-08377,DBLP:conf/emnlp/ZhaoLZHCHZ24,DBLP:conf/acl/0025FDGY0CCC024}, we use the retrieved passages $P$ as the reference system for rational evidence $e$, encouraging it to be significantly shorter than $P$ while preserving key clues.
Conversely, $e$ serves as the reference for the rationale $r$, encouraging it to be comprehensive enough to capture all key clues in the retrieved passages.
Based on these principles, we define length rewards for both the rationale and rational evidence, denoted as $R^{\text{len}}_{r}$ and $R^{\text{len}}_{e}$, respectively, where $R^{len}_{r}, R^{len}_{e} \in (0,1]$.
Detailed formulations and analyses of the length rewards are provided in Appendices~\ref{app:length_def} and \ref{app:length_ana}.

\smallsection{Format Reward}
It focuses on whether the response obeys the defined format. 
Specifically, the model’s evidence reasoning, evidence extraction, as well as final output should be enclosed within the \textbf{\textcolor[RGB]{68,114,196}{<reason>}} ... \textbf{\textcolor[RGB]{68,114,196}{</reason>}},  \textbf{\textcolor[RGB]{0,100,0}{<extract>}} ... \textbf{\textcolor[RGB]{0,100,0}{</extract>}}, and \textbf{\textcolor[RGB]{127,100,0}{<answer>}} ... \textbf{\textcolor[RGB]{127,100,0}{</answer>}} tags, respectively. 
Therefore, the format reward $R^{fmt}$ is assigned a value of 1 only if responses fully adhere to the formatting requirements; otherwise, it is set to 0.
\smallsection{Putting them together} After obtaining the answer, length, and format rewards, we compute the final reward via a linear weighted sum, as complex weighting is not the focus of this work, and a linear one generally leads to satisfactory performance:
\begin{equation}
    R^{final} = \lambda_{1} R^{ans} + \lambda_{2} R^{len} + \lambda_{3} R^{fmt},
\end{equation}
where $R^{ans}$ is the average answer reward; $R^{len}$ is the average length reward; $\lambda_*$ is the weighting coefficient; $R^{final}$ is the final reward used in RL.
\begin{table*}[t]
  \renewcommand\arraystretch{1.08}
  \tabcolsep=0.205cm
  \centering
  \footnotesize
  \begin{tabular}{cc|cc|ccccc|cc}
    \toprule
    \multicolumn{1}{c}{\multirow{2}{*}{\centering \bf Datasets}} 
    & \multicolumn{1}{c|}{\multirow{2}{*}{\centering \bf Metrics}} 
    & \multicolumn{2}{c|}{\centering \bf WR}
    & \multicolumn{5}{c|}{\centering \bf VAR} 
    & \multicolumn{2}{c}{\centering \bf RAR} \\
    \cline{3-11}
    &  & \multirow{1}{*}[-0.3ex]{\centering Zero} & \multirow{1}{*}[-0.3ex]{\centering Full} & \multirow{1}{*}[-0.3ex]{\centering SeleCtx}  &  \multirow{1}{*}[-0.3ex]{\centering LLMLingua-2} &  \multirow{1}{*}[-0.3ex]{\centering \textsc{Recomp}} &  \multirow{1}{*}[-0.3ex]{\centering FilCo} & \multirow{1}{*}[-0.3ex]{\centering SEER} & \multirow{1}{*}[-0.3ex]{\centering CoT} & \multirow{1}{*}[-0.3ex]{\centering {\ours{}}}\\ 
    \hline
   \rowcolor[HTML]{D3D3D3} 
   \multicolumn{11}{c}{\textbf{\text{1.5B size model}}} \\
     \cellcolor{white!20} &  {EM} & 13.74 & \textbf{41.97} & 24.68 & 29.36 & 37.40 &  36.62 & 36.93 &  37.70 & \underline{41.14} \\
    \cellcolor{white!20} &  {F1} & 53.16 & \underline{70.07} & 59.52 & 62.82 & 67.18 & 66.63 & 67.11 & 68.09 & \textbf{70.77} \\
     \cellcolor{white!20} \multirow{-3}{*} {\centering \textbf{NQ}}  &  {CR} & - & 1.0x & 3.44x & 4.51x & 5.43x & \underline{16.3x} & 13.2x & 4.56x & \textbf{38.1x} \\
    \rowcolor{gray!12}
     &  {EM} & 29.10 & \textbf{57.02} & 45.13 & 48.67 & 56.56 &  54.06 & 54.57 &  54.29 & \underline{56.84} \\
    \rowcolor{gray!12}
     &  {F1} & 64.62 & \underline{80.13} & 73.40 & 75.29 & 79.85 & 78.56 & 78.81 & 79.56 & \textbf{80.85} \\
    \rowcolor{gray!12}
      \multirow{-3}{*} {\centering \textbf{TQA}$^{\dag}$}  &  {CR} & - & 1.0x & 3.38x & 4.52x & 5.35x & 8.55x & \underline{10.3x} & 5.02x & \textbf{38.8x} \\
    \cellcolor{white!20} &  {EM} & 12.36 & 19.20 & 15.30 & 16.64 & 18.52 & 18.18  & 18.60 & \underline{19.52}  & \textbf{20.46} \\
    \cellcolor{white!20} &  {F1} & 48.52 & 53.04 & 49.65 & 51.26 & 52.92 & 52.15 & 52.79 & \underline{53.43} & \textbf{54.20} \\
     \cellcolor{white!20} \multirow{-3}{*} {\centering \textbf{HotpotQA}$^{\dag}$}  &  {CR} & - & 1.0x & 3.40x & 4.52x & 5.44x & \underline{18.3x} & 15.5x & 4.17x & \textbf{33.0x} \\
    \hline
    \rowcolor[HTML]{D3D3D3}
   \multicolumn{11}{c}{\textbf{\text{7B size model}}} \\
     \cellcolor{white!20} &  {EM} & 25.04 & \textbf{48.78} & 33.91 & 36.51 & 43.77 & 44.79  & 45.01 &  44.49 & \underline{46.95} \\
    \cellcolor{white!20} &  {F1} & 60.88 & \textbf{74.40} & 65.71 & 67.77 & 71.61 & 72.30 & 72.76 & 73.09 & \underline{73.99} \\
     \cellcolor{white!20} \multirow{-3}{*} {\centering \textbf{NQ}}  &  {CR} & - & 1.0x & 3.44x & 4.51x & 5.43x & \underline{15.4x} & 12.4x & 3.36x & \textbf{17.9x}\\
    \rowcolor{gray!12}
     &  {EM} & 47.31 & \textbf{65.34} & 58.72 & 60.50 & 64.40 & 63.46  & 64.20 &  63.91 & \underline{64.76} \\
    \rowcolor{gray!12}
     &  {F1} & 74.36 & \textbf{84.83} & 80.95 & 81.94 & 84.12 & 83.66 & 84.14 & 84.45 & \underline{84.74} \\
    \rowcolor{gray!12}
      \multirow{-3}{*} {\centering \textbf{TQA}$^{\dag}$}  &  {CR} & - & 1.0x & 3.38x & 4.52x & 5.35x & 7.77x & \underline{9.70x} & 3.33x & \textbf{17.6x} \\
    \cellcolor{white!20} &  {EM} & 17.95 & 25.82 & 21.27 & 23.45 & 24.84 & 25.27  & 25.81 &  \underline{27.25} & \textbf{28.02} \\
    \cellcolor{white!20} &  {F1} & 53.07 & 58.50 & 54.90 & 56.65 & 58.02 & 58.15 & 58.63 & \underline{59.84} & \textbf{60.30} \\
     \cellcolor{white!20} \multirow{-3}{*} {\centering \textbf{HotpotQA}$^{\dag}$}  &  {CR} & - & 1.0x & 3.40x & 4.52x & 5.44x & \textbf{17.5x} & 14.3x & 3.32x & \underline{16.5x} \\
    \bottomrule
  \end{tabular}
  \caption{Overall performance comparison, where the best results are \textbf{boldfaced} and the second-best results are \underline{underlined}. 
  ${\dag}$ denotes out-of-domain (OOD) datasets for our \ours{}, since it is only trained on the NQ dataset.
  The improvements of \ours{} over other VAR and RAR methods are statistically significant with $p$-value < 0.05.}
  \label{tab:overall}
\end{table*}

\subsection{Online Policy Optimization}
\label{sec:online}
Having obtained the final rewards for each response, we optimize \ours{} using the policy optimization algorithm GRPO~\cite{DBLP:journals/corr/abs-2402-03300,DBLP:journals/corr/abs-2501-12948}, thereby incentivizing rational evidence extraction capability in it via reinforcement learning. 
Specifically, for each input question $q$, GRPO first samples a group of responses $\mathcal{Y} = \{y_1, y_2, ..., y_G\}$, where $G$ is the group size, and each response comprises the rationale $r$, rational evidence $e$, and three conditional outputs $o_r$, $o_e$, and $o_f$ (\S\ref{sec:obtain}).
Subsequently, GRPO evaluates these responses using the verifiable reward functions (\S\ref{sec:modeling}) and obtains final rewards for each response, denoted as $\mathcal{R} = \{R_1, R_2, ..., R_G\}$, where we omit the superscript ``${final}$'' for brevity.
Unlike PPO \cite{DBLP:journals/corr/SchulmanWDRK17}, GRPO directly compares the final rewards of candidate responses within the same group without requiring an additional critic model.
The advantage of the $i$-th response is determined by normalizing its reward $R_i$ using the mean and the standard deviation of rewards in $\mathcal{R}$, which can be formulated as follows:
\begin{equation}
\label{equ:adv}
    A_i = \frac{R_i-\mathrm{mean}(\mathcal{R})}{\mathrm{std}(\mathcal{R})},
\end{equation}
where $\mathrm{mean}(\cdot)$ and $\mathrm{std}(\cdot)$ compute the average and standard deviation of the input set, respectively.
However, GRPO's group normalization may overly magnify minor numerical fluctuations.
For example, considering $\mathcal{R} = \{0.49, 0.51\}$, the $\mathrm{mean}(R)$ is $0.5$, the $\mathrm{std}(R)$ is $0.01$, and the resulting advantages are $\{-1.0,1.0\}$, which actually magnifies minor fluctuations.
To address this issue, we propose clipping $\mathrm{std}(R)$ such that it is at least $\epsilon_{std}$, ensuring the denominator does not become too small:
\begin{equation}
\label{equ:clip_adv}
    \tilde{A}_i = \frac{R_i-\mathrm{mean}(\mathcal{R})}{\mathrm{clip\_std}(\mathcal{R})},
\end{equation}
where $\mathrm{clip\_std}(\mathcal{R}) = \mathrm{max}(\mathrm{std}(R), \epsilon_{std})$; $\epsilon_{std}$ is a clipping value.
After obtaining the refined advantages $\{\tilde{A}_1, \tilde{A}_2, ..., \tilde{A}_G\}$, we optimize the current policy model $\pi_{\theta}$ (the evidence extractor $\mathcal{M}_\mathcal{E}(\cdot)$) via maximizing the following objective function:
\begin{equation}
\small
\begin{aligned}
J_{\mathrm{GRPO}}(\theta) 
&= \mathbb{E}_{x \sim \mathcal{D},\, \{y_i\}_{i=1}^G \sim \pi_{\theta_{\mathrm{old}}}(\cdot \mid x)} \\
&\Biggl[
  \frac{1}{G} \sum_{i=1}^G
  \min\!\Bigl(
    \frac{\pi_{\theta}(\tilde{y}_i \mid x)}{\pi_{\theta_{\mathrm{old}}}(\tilde{y}_i \mid x)}\,\tilde{A}_i,\, \\
    &\mathrm{clip}\!\Bigl(
      \frac{\pi_{\theta}(\tilde{y}_i \mid x)}{\pi_{\theta_{\mathrm{old}}}(\tilde{y}_i \mid x)},
      1-\varepsilon,\,
      1+\varepsilon
    \Bigr)
    \tilde{A}_i
  \Bigr)  \\
  &-\,\beta\,D_{\mathrm{KL}}\bigl(\pi_{\theta}\,\big\|\,\pi_{\mathrm{ref}}\bigr)
\Biggr],
\end{aligned}
\label{equ:grpo}
\end{equation}
where $x=\{q, P\}$ denotes input samples drawn from the dataset $\mathcal{D}$; $y$ is the model's response consisting of $\{r,e,o_r,o_e,o_f\}$, sampled from the old policy $\pi_{\theta_{\mathrm{old}}}$; 
$\tilde{y}$ is a response consisting of $\{r,e,o_f\}$, where $o_f$ is an output conditioned on full context. This makes $\tilde{y}$ self-contained, and it is therefore used for training;
$\epsilon$ and $\beta$ are the PPO clipping hyperparameters and KL weighting coefficient, respectively; $\pi_{\mathrm{ref}}$ represents the reference policy.
\section{Experiment}
In this section, we conduct extensive experiments on five benchmark datasets to answer the following Research Questions (\textbf{RQs}):
\begin{itemize}[leftmargin=*]
    \item \textbf{RQ1:} How does rational evidence perform compared to vanilla evidence?
    \item \textbf{RQ2:} How do the properties of rational evidence vary with the RL training process?
    \item \textbf{RQ3:} How do different parts of the answer rewards contribute to the final model performance?
    \item \textbf{RQ4:} How does the inference efficiency of \ours{} compare to different types of methods?
    \item \textbf{RQ5:} Can rational evidence perform robustly against retrieval noise?
    \item \textbf{RQ6:} How does rational evidence perform on challenging multi-hop QA? (see~Appendix \ref{app:multi_qa_bench})
    \item \textbf{RQ7:} Can rational evidence be effective for agentic RAG? (see Appendices~\ref{app:poten_evi_agent} and \ref{app:study_on_avg_search})
\end{itemize}

\begin{table}[ht]
  \centering
  \footnotesize
  \tabcolsep=0.18cm
  \renewcommand\arraystretch{1.1}
  \begin{tabular}{l|ccc}
    \toprule
    \multicolumn{1}{l|}{\textbf{Dataset}} & 
    \multicolumn{1}{c}{\textbf{\#Train}} & \multicolumn{1}{c}{\textbf{\#Dev}} & \multicolumn{1}{c}{\textbf{\#Test}}\\
    \hline
    NQ~\cite{DBLP:journals/tacl/KwiatkowskiPRCP19}   & 79.1k & 8.7k & 3.6k \\
    TQA~\cite{DBLP:conf/acl/JoshiCWZ17}  & 78.7k & 8.8k & 11.3k \\
    HotoptQA~\cite{DBLP:conf/emnlp/Yang0ZBCSM18}  & 88.9k & 5.6k & 5.6k \\
    2WikiQA~\cite{DBLP:conf/coling/HoNSA20}  & 167.4k & 12.5k & 12.5k \\
    Musique~\cite{DBLP:journals/tacl/TrivediBKS22}  & 19.9k & 2.4k & 2.4k \\
    \bottomrule
  \end{tabular}
  \caption{\centering{Statistics of the datasets.}}
  \label{tab:dataset}
  \vspace{-0.15cm}
\end{table}
\begin{figure*}[t]
    \centering  
     \subfigure[Answer reward dynamics on 1.5B model.]{
        \includegraphics[width=.385\linewidth]{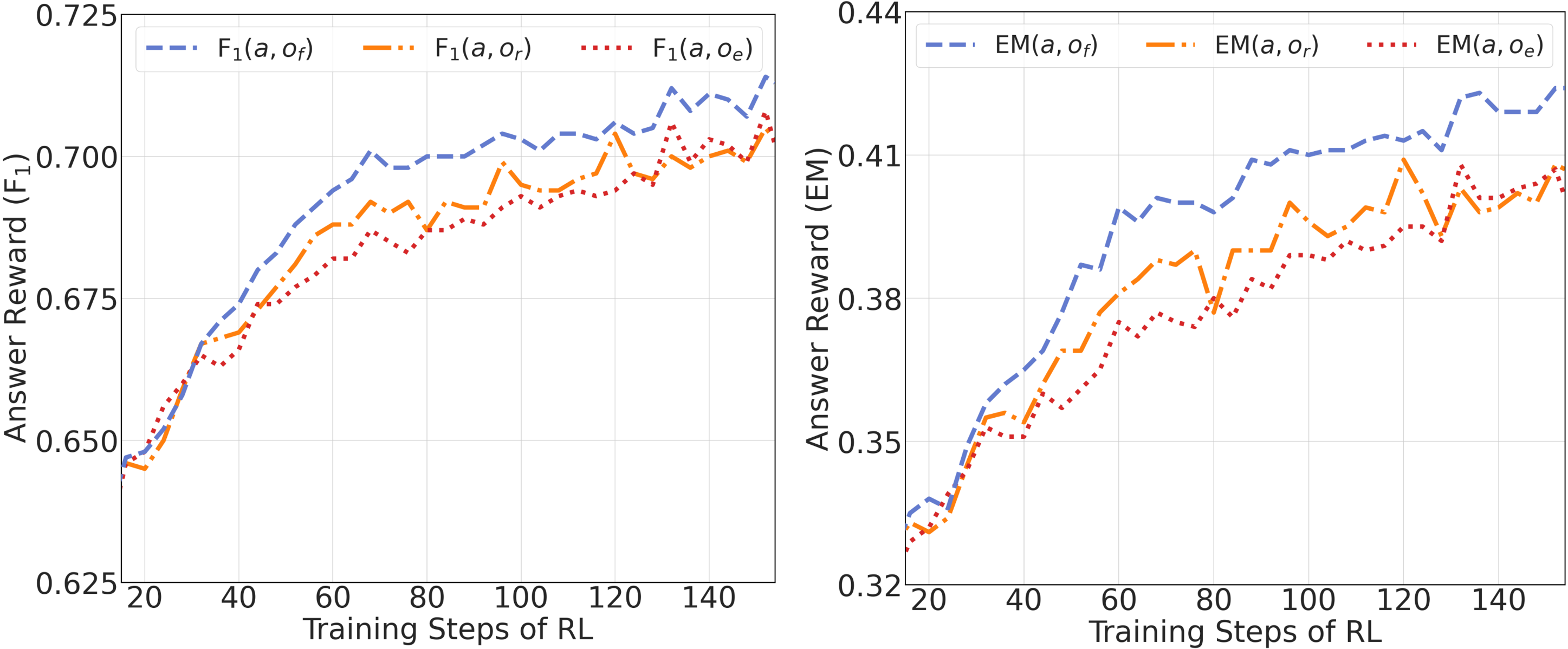}\label{fig:nq_answer_reward_1_5b}}
    \subfigure[Answer reward dynamics on 7B model.]{
        \includegraphics[width=.385\linewidth]{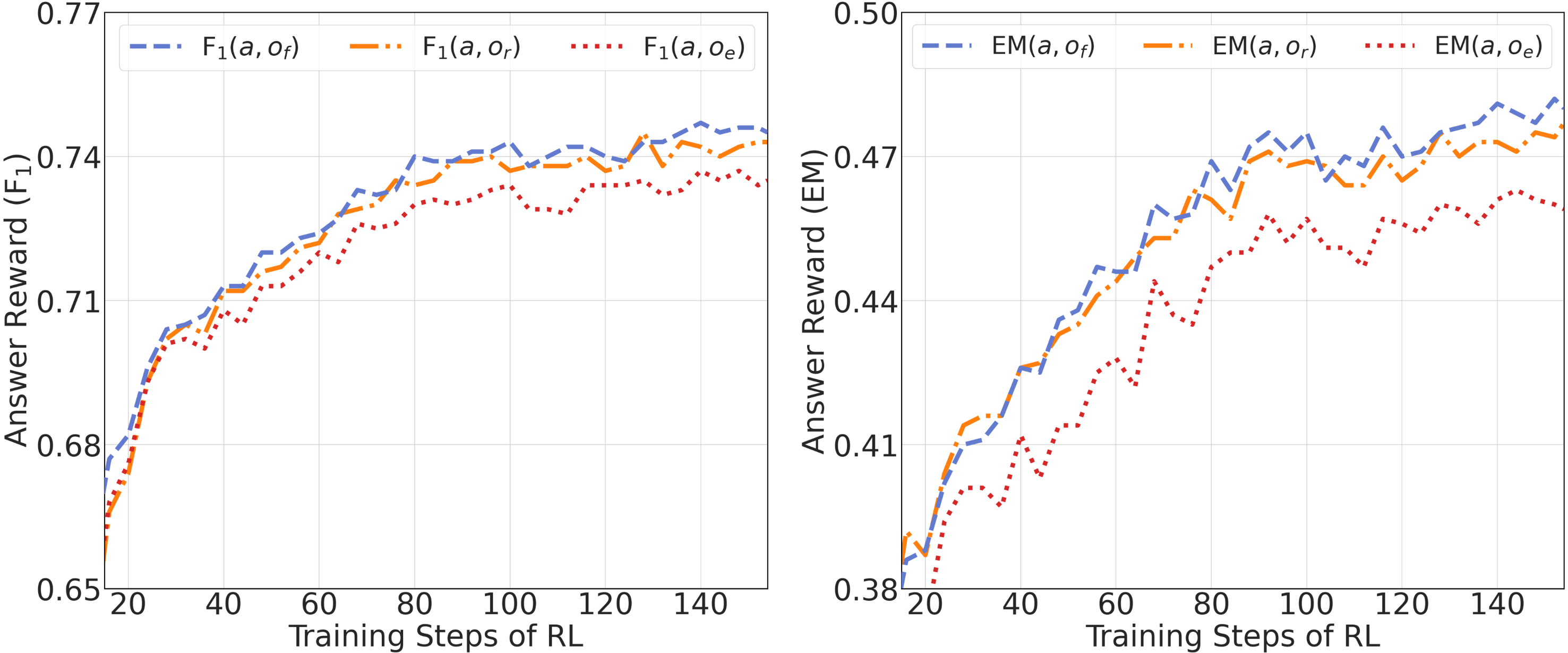}\label{fig:nq_answer_reward_7b}}
    \subfigure[Length dynamics.]{
        \includegraphics[width=.185\linewidth]{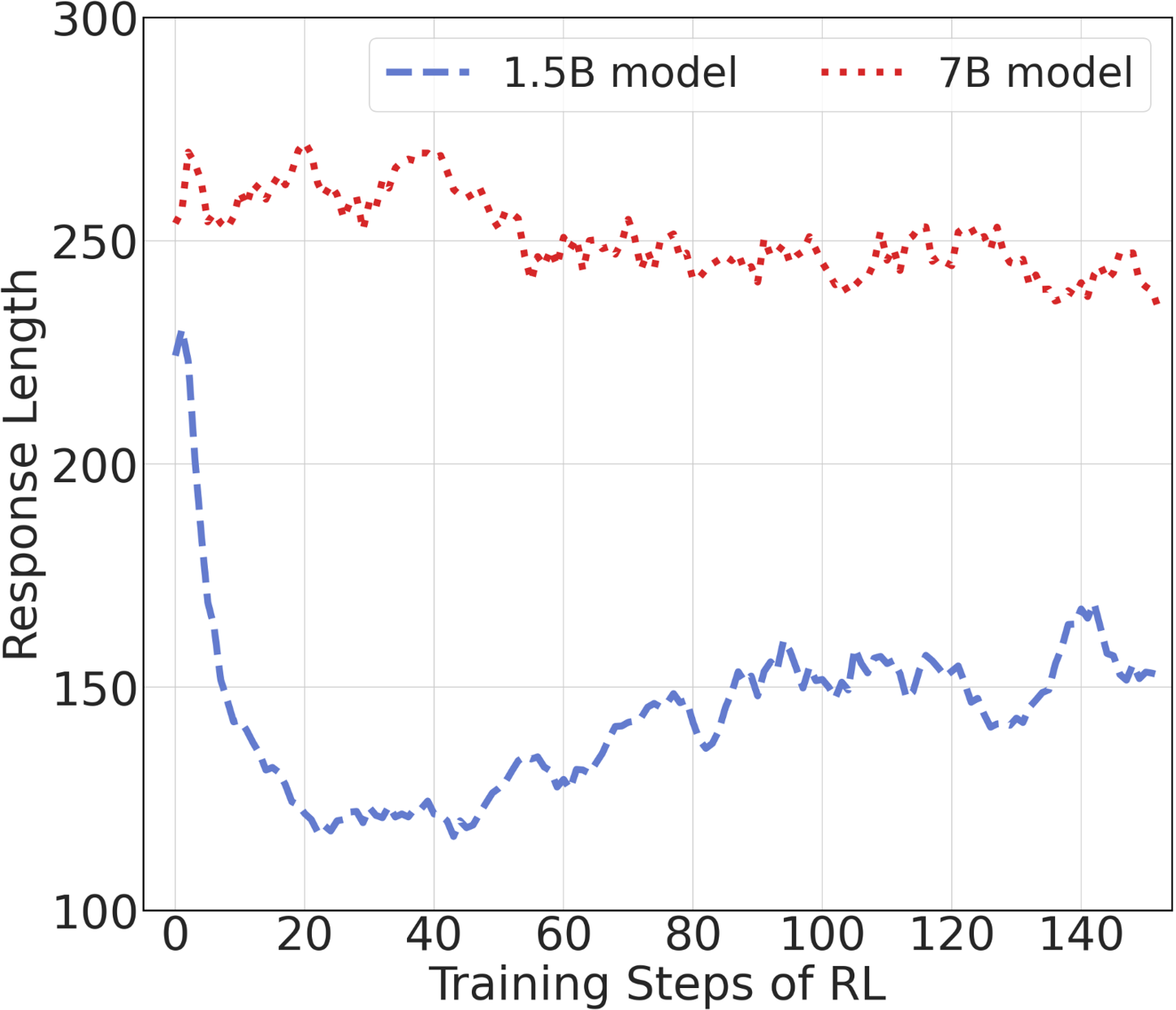}\label{fig:nq_response_length}}
    \caption{Training dynamics \textit{w.r.t.} answer reward and response length (including `reason', `extract', and `answer').}
    \label{fig:training_dynamics}
\end{figure*}
\begin{table*}[t]
  \centering
  \small
  \tabcolsep=0.16cm
  \renewcommand\arraystretch{1.1}
  \begin{tabular}{l|ccc|cc|c}
    \toprule
    \multicolumn{1}{l|}{\textbf{Models}} & 
    \multicolumn{1}{c}{Rationale} & \multicolumn{1}{c}{Evidence} & \multicolumn{1}{c}{Full Ctx} & \multicolumn{1}{|c}{Evidence+Rationale}  & \multicolumn{1}{c}{Evidence+Full Ctx}  & \multicolumn{1}{|c}{Evidence+Rationale+Full Ctx}\\
    \hline 
    {NQ} & 39.89 & 40.27 & 39.53 & 40.02 & \underline{40.44} & \textbf{41.14} \\
    {TQA} & 56.97 & \textbf{57.38} & 56.17 & \underline{57.31} & 57.26 & 56.84 \\
    {HotpotQA} & 20.30  & 19.73 & 20.13 & \underline{20.34} & 20.23 & \textbf{20.46}\\
    \hline
    {Average} & 39.05 & 39.13 & 38.61 & 39.22 & \underline{39.31} & \textbf{39.48}\\
    \bottomrule
  \end{tabular}
  \caption{Performance comparison (exact match) of \ours{} trained on different combinations of answer rewards.}
  \label{tab:abla}
\end{table*}
\subsection{Experimental Setup}
We experiment on five benchmark QA datasets, including Natural Questions (NQ)~\cite{DBLP:journals/tacl/KwiatkowskiPRCP19}, TriviaQA (TQA)~\cite{DBLP:conf/acl/JoshiCWZ17}, HotpotQA~\cite{DBLP:conf/emnlp/Yang0ZBCSM18}, 2WikiQA~\cite{DBLP:conf/coling/HoNSA20}, and MuSiQue~\cite{DBLP:journals/tacl/TrivediBKS22}, where the first two are open-domain QA, and the last three are multi-hop QA.
The detailed statistics of these datasets are presented in Table \ref{tab:dataset}.
We use the Exact Match ({EM}$\uparrow$) and unigram {$\mathrm{F}_1$$\uparrow$} to evaluate QA performance, and use the Compression Ratio (CR$\downarrow$) to measure efficiency.
We employ Qwen2.5-1.5B-Instruct and Qwen2.5-7B-Instruct~\cite{qwen2025qwen25technicalreport} as the initial models and train these two models to extract rational evidence via full parameter fine-tuning for 1 epoch on the NQ dataset. 
For retrieval, we use Dense Passage Retriever \cite{DBLP:conf/emnlp/KarpukhinOMLWEC20} to fetch top-5 passages from all Wiki passages.
For comparison, we compare \ours{} with the three groups of competitive baselines: \textbf{(1) Without Refinement (WR):} \textbf{(\rmnum{1})} Zero-shot (Zero); \textbf{(\rmnum{2})} Full Passage (Full);
\textbf{(2) Vanilla Refinement (VAR):} \textbf{(\rmnum{1})} Select Context (SeleCtx) \cite{DBLP:conf/emnlp/0001DGL23}; 
\textbf{(\rmnum{2})} LLMLingua-2 \cite{DBLP:conf/acl/PanWJXLZLR0LZQ024}; \textbf{(\rmnum{3})} \textsc{Recomp} \cite{DBLP:conf/iclr/XuSC24}; 
\textbf{(\rmnum{4})} \textsc{FilCo} \cite{DBLP:journals/corr/abs-2311-08377}; 
\textbf{(\rmnum{5})} SEER \cite{DBLP:conf/emnlp/ZhaoLZHCHZ24};
\textbf{(3) Rational Refinement (RAR):} \textbf{(\rmnum{1})} Chain-of-Thought (CoT) \cite{DBLP:conf/nips/Wei0SBIXCLZ22}. 
Detailed experimental settings and prompts can be found in \textbf{Appendices}~\ref{app:experi_set} and \ref{app:prompt}.
\subsection{Main Comparison (RQ1)}
The overall comparison results are presented in Table \ref{tab:overall}. From the results, we have the following observations.
\textbf{(1)} In all cases (18/18), \ours{} achieves the best or second-best results, indicating the superiority of supplementing RAG with rational evidence.
\textbf{(2)} Surprisingly, directly applying \ours{} trained on NQ and tested on OOD datasets also yields impressive performance, indicating that RL can endow \ours{} with superior generalization.
\textbf{(3)} Compared to `Full', \ours{} achieves an extremely high compression ratio (\textit{e.g.,} 38.1x with 1.5B size model on NQ), while its performance is very close to or even surpasses that of `Full'.
In contrast, although competitive baselines have relatively high compression ratios, their compression is accompanied by large performance degradation.
\textbf{(4)} By comparing the 1.5B  and 7B variants of \ours{}, we find that the 7B one tends to extract more informative evidence than the 1.5B one.
\textbf{(5)} \ours{} considerably outperforms VAR methods in nearly all cases and provides more compact evidence, showing the necessity of evidence reasoning. 
\textbf{(6)} In HotpotQA, RAR methods largely outperform VAR ones and even surpass `Full', indicating that rational refinement is important to multi-hop QA.
These observations fully manifest the effectiveness and efficiency of rational evidence.
\subsection{Training Dynamics (RQ2)}
The RL training dynamics \textit{w.r.t.} the answer rewards and response length are shown in Figure~\ref{fig:training_dynamics}. 
The results show that the answer rewards of full context ($o_f$), rationale ($o_r$), and rational evidence ($o_e$) are consistently improved during the process of RL.
Unsurprisingly, the answer reward of full context generally performs best.
Specifically, on the 1.5B model, rationale and rational evidence perform very closely, while on the 7B model, rational evidence performs slightly worse than rationale.
We think the main reason is that the 7B model can infer answers based on implicit clues, whereas rational evidence may omit some of them.
For the dynamics of response length, the response length of the 1.5B model decreases rapidly at the beginning of training and then increases slowly, while that of the 7B model decreases slowly all the time.
We think the main reason is that a moderate length is beneficial for improving answer rewards, as an overlong reasoning may confuse answer generation, and a too short reasoning may omit key clues.
Therefore, as training continues, the response lengths of the 1.5B model and the 7B model tend to converge.
In all, the above observations provide several useful insights for reward modeling as well as designing.
\subsection{Ablation Study (RQ3)}
To evaluate the impact of different answer rewards on model performance, we train the 1.5B variant of \ours{} models using different combinations of answer rewards with average weighting.
As shown in Table~\ref{tab:abla}, among models trained on a single answer reward, training with evidence's answer reward $R^{ans}_e$ achieves the highest average performance, indicating that the quality of evidence is the most important.
For two-answer reward combinations, `Evidence+Full Ctx' slightly outperforms `Evidence+Rationale', with significant gains on NQ, indicating that the optimization of `Full Ctx' matters.
Notably, using all answer rewards yields the best performance on NQ and HotpotQA, and the highest average score, indicating the benefits of multi-faceted rewarding.
The results manifest the contribution and necessity of each answer reward.
\begin{table}[t]
  \centering
  \footnotesize
  \tabcolsep=0.3cm
  \renewcommand\arraystretch{1.1}
  \begin{tabular}{l|ccc|c}
    \toprule
    \multicolumn{1}{l|}{\textbf{Models}} & 
    \multicolumn{1}{c}{NQ} & \multicolumn{1}{c}{TQA} & \multicolumn{1}{c}{HotpotQA} & \multicolumn{1}{|c}{Avg} \\
    \hline 
    \textsc{FilCo}     & 0.64 & 0.82 & \underline{0.59} & 0.68\\
    CoT   & \underline{0.55} & \underline{0.70} & 0.71 & \underline{0.65} \\
    \ours{}  & \textbf{0.35} & \textbf{0.41} & \textbf{0.43} & \textbf{0.40} \\
    \bottomrule
  \end{tabular}
  \caption{Inference latency (seconds/query) on 1.5B.}
  \label{tab:latency}
\end{table}
\subsection{Inference Efficiency (RQ4)}
Table~\ref{tab:latency} presents the inference latency of \ours{} compared to \textsc{FilCo} (a representative VAR method) and CoT (a representative RAR method) on the 1.5B model.
All experiments are conducted on an A800 GPU, with the batch size and max new tokens set to 64 and 768, respectively.
The results show that the latency of evidence extraction using \ours{} is considerably shorter than \textsc{FilCo} and CoT.
Surprisingly, the average latency of \textsc{FilCo} is slightly longer than that of CoT, while FilCo generates fewer new tokens on average.
Our analysis shows that the length's std of evidence extracted by \textsc{FilCo} is largely higher than that of COT, where the std of \textsc{FilCo} on NQ, TQA, and HotpotQA is 107.5, 136.1, and 81.7, while that of CoT is 67.9, 61.6, and 63.9.
The unstable evidence length of \textsc{FilCo} disrupts efficient batching.
The above results show the efficiency of \ours{} during deployment.
\subsection{Robustness Analysis (RQ5)}
\begin{figure}[t]
    \centering  
     \subfigure[NQ.]{
        \includegraphics[width=.48\linewidth]{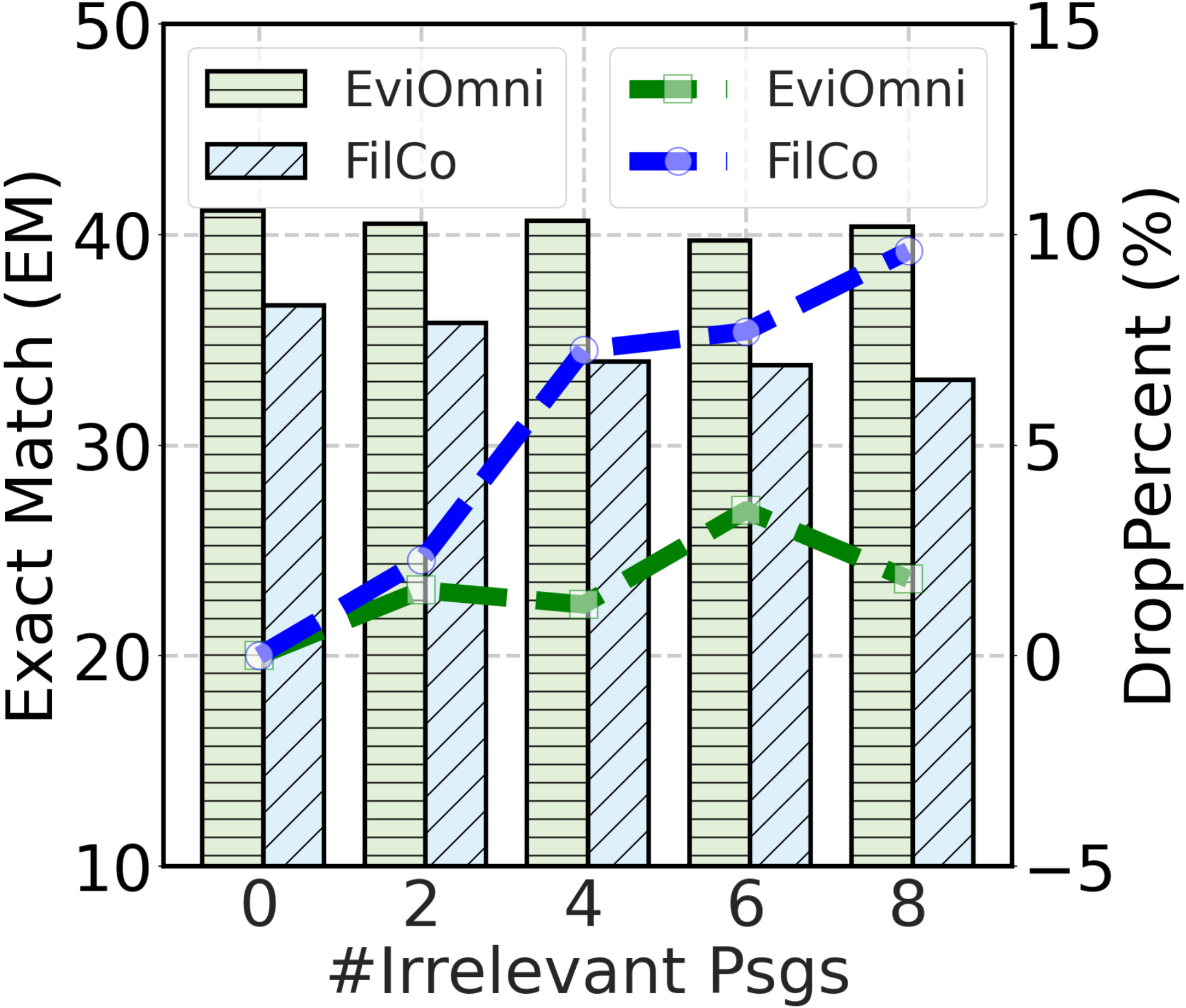}\label{fig:nq_noise_em}}
    \subfigure[TQA.]{
        \includegraphics[width=.48\linewidth]{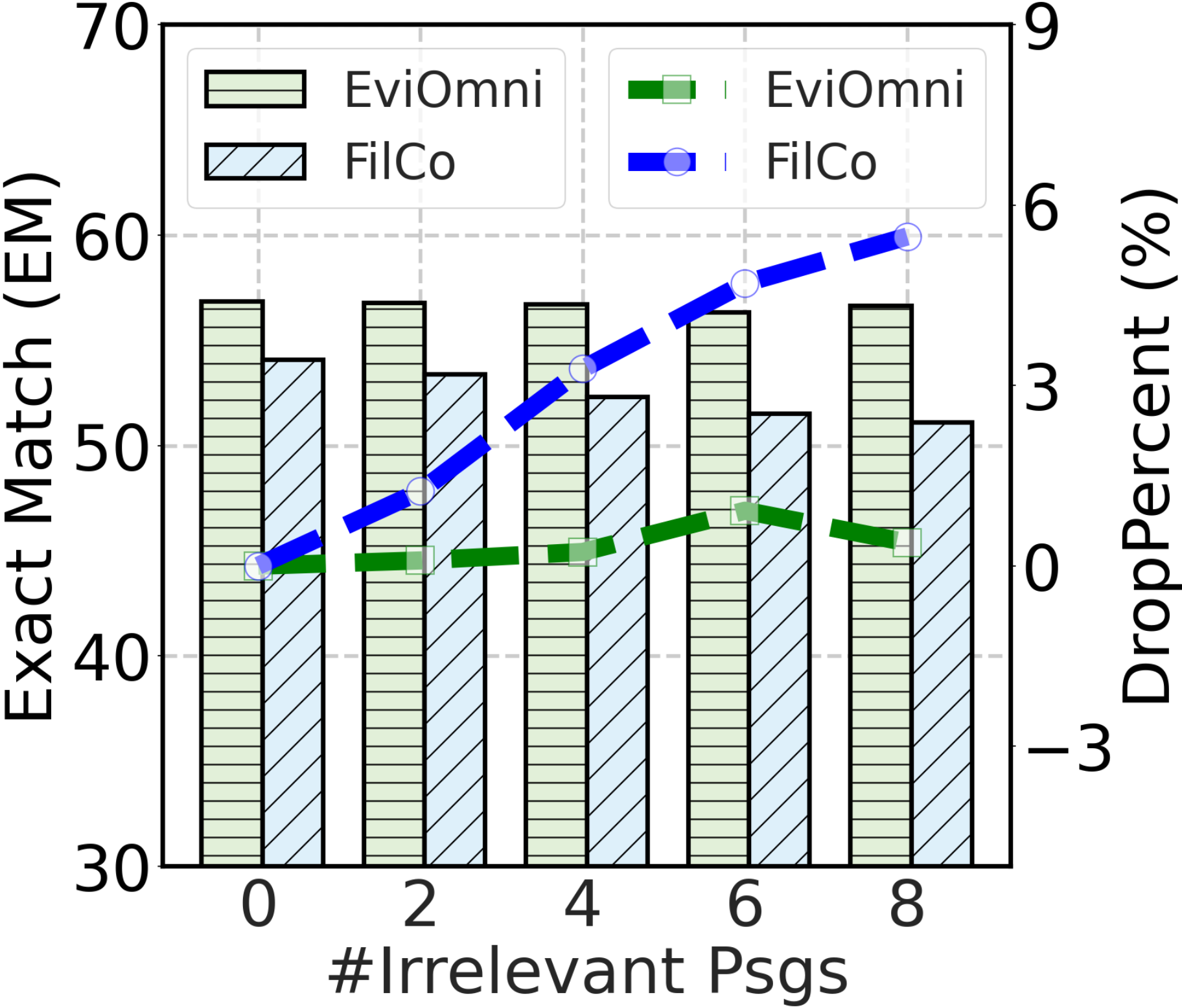}\label{fig:tqa_noise_em}}
    \caption{Performance comparisons \textit{w.r.t.} data noise, where Psgs is the abbreviation of passages.}
    \label{fig:robust}
\end{figure}

In real-world applications, RAG systems often suffer from data noise caused by imperfect retrieval. 
To simulate this scenario, we randomly sample a certain number (\textit{i.e.,} 0, 2, 4, 6, and 8) of irrelevant passages for each test query.
Then, each query is paired with 5 retrieved relevant passages and sampled irrelevant passages.
We experiment on the 1.5B model, and the results are presented in Figure~\ref{fig:robust}.
The results show that adding noise considerably degrades the performance of \textsc{FilCo}, while the performance drop of \ours{} is relatively small, where the green line is always below the blue one.
Particularly, \ours{} with 8 irrelevant passages (psgs) outperforms \textsc{FilCo} without noise.
This fully manifests the robustness of \ours{} against noise.
\section{Related Works}
\subsection{Retrieval-Augmented Generation}
RAG prevails in LLMs, enhancing LLMs with external non-parametric knowledge to remedy their outdated, incorrect, or incomplete internal parametric knowledge~\cite{DBLP:conf/nips/LewisPPPKGKLYR020,DBLP:conf/emnlp/WuZHMS022,DBLP:journals/corr/abs-2002-08909,DBLP:journals/jmlr/IzacardLLHPSDJRG23,DBLP:conf/iclr/AsaiWWSH24,DBLP:journals/tacl/ChenZCWLZZ24,DBLP:journals/tmm/RaoDQFLST23,DBLP:conf/acl/RaoLLKLJCYZ25}.
The pioneering attempts \cite{DBLP:conf/nips/LewisPPPKGKLYR020,DBLP:journals/corr/abs-2002-08909} demonstrated that augmenting the input context of LLMs with retrieved passages yields significant improvements, especially in knowledge-intensive tasks.
While prior works usually retrieve straightforwardly at the beginning, dynamic RAG
\cite{DBLP:conf/iclr/AsaiWWSH24,DBLP:conf/naacl/JeongBCHP24,DBLP:conf/emnlp/JiangXGSLDYCN23} proposed to adaptively retrieve passages based on the demand of generation and the complexity of the query.
Recently, agentic RAG~\cite{DBLP:journals/corr/abs-2506-18959} proposed to interleave retrieval and reasoning to tackle complex queries.
In particular, this kind of work~\cite{DBLP:journals/corr/abs-2503-05592,DBLP:journals/corr/abs-2503-09516,DBLP:conf/acl/TrivediBKS23} focuses on prompting or fine-tuning LLMs as search agents that interact with search tools to fetch external knowledge.
\subsection{RAG with Context Compression}
Traditional RAG systems typically concatenate all retrieved passages as the context of LLMs~\cite{DBLP:conf/nips/LewisPPPKGKLYR020,zhao2026kalmembeddingv}.
However, this may introduce data noise and computational overhead due to imperfect retrieval and overlong context~\cite{DBLP:journals/corr/abs-2311-08377}.
Recently, many works have attempted to compress context and retain important clues, mainly including two categories: 
\textbf{(1)} Reranking methods, which rerank retrieved passages and retain top-ranked passages \cite{DBLP:journals/corr/abs-2407-03627,DBLP:journals/corr/abs-2501-04695,DBLP:conf/acl/MaoHLSGHC21,chen2025beyond};
\textbf{(2)} Summarization methods, which extract relevant information from retrieved passages and consolidate them into a narrative~\cite{DBLP:journals/corr/abs-2311-08377,DBLP:conf/emnlp/ZhaoLZHCHZ24,DBLP:conf/acl/0025FDGY0CCC024,DBLP:conf/iclr/XuSC24}.
Despite their effectiveness, existing works overlook evidence reasoning, thereby risking filtering out key clues as well as struggling with generalization and robustness.
\section{Conclusion}
In this paper, we first unveil the limitations of vanilla evidence, which risks filtering out crucial clues, and then empirically validate the superiority of rational evidence in substantially retaining key clues.
To this end, we propose a rational evidence extraction learning framework, \ours{}, which first reasons and then extracts, is optimized by reinforcement learning with verifiable rewards, incorporates knowledge token masking to avoid information leakage, and ultimately produces compact, high-quality evidence.
Extensive experiments across both traditional and agentic RAG scenarios demonstrate the superiority of \ours{} in end-to-end performance, out-of-domain generalization, inference efficiency, and robustness to retrieval noise.
\section*{Limitations}
Despite our innovations and improvements, \ours{} still has certain limitations, especially cascaded generation between rationale and rational evidence.
This indicates that the rationale must be generated first, followed by the rational evidence, which increases the inference latency to some extent.
Although the results in Table~\ref{tab:latency} show that the inference efficiency of \ours{} is superior to that of the baselines, we believe that further optimization of reasoning efficiency is necessary to achieve higher inference speed. 
One promising direction is reasoning when necessary \cite{DBLP:journals/corr/abs-2505-14631}.
For example, for simple queries, the model could directly generate rational evidence without reasoning, thereby reducing redundant computation. 
We leave this adaptive framework for future research.
\section*{Acknowledgements}
This work is supported by the National Natural Science Foundation of China (Grant No. 62336008), Shenzhen Basic Research Program (Grant No. JCYJ20241202123503005), and Shenzhen Basic Research Program (Grant No. JCYJ20240813105111016).
We sincerely thank all reviewers for their insightful comments and valuable suggestions, which have significantly improved our work.

\bibliography{custom}

\appendix

\section{Experimental Setup}
\label{app:experi_set}
\subsection{Datasets and Metrics} 
We experiment on five knowledge-intensive benchmark QA datasets, including Natural Questions (NQ)~\cite{DBLP:journals/tacl/KwiatkowskiPRCP19}, TriviaQA (TQA)~\cite{DBLP:conf/acl/JoshiCWZ17}, HotpotQA~\cite{DBLP:conf/emnlp/Yang0ZBCSM18}, 2WikiQA~\cite{DBLP:conf/coling/HoNSA20}, and MuSiQue~\cite{DBLP:journals/tacl/TrivediBKS22}, where the first two are open-domain QA, and the last three are multi-hop QA.
The detailed statistics of these five datasets are provided in Table \ref{tab:dataset}, where the test set for HotpotQA is unavailable; therefore, we use its development set as a substitute for the test set.
Additionally, since the test set of 2WikiQA or MuSiQue is not annotated, we use their development sets for evaluation.
For evaluation, we adopt the Exact Match ({EM}) and unigram {$\mathrm{F}_1$} score to evaluate QA performance.
EM examines exact correctness while $\mathrm{F}_1$ calculates the degree of lexical overlap, offering a more fine-grained view of how well the prediction aligns with the golden answer.
To measure the improvement of the computational efficiency, we employ the Compression Ratio (CR) following the previous works \cite{DBLP:journals/corr/abs-2412-12559,DBLP:conf/acl/PanWJXLZLR0LZQ024}, where CR computes the ratio of the total length of the retrieval passages to the length of the extracted evidence: 
\begin{equation}
\mathrm{CR} = \frac{\sum_{p\in P} \mathrm{L}_p}{\mathrm{L}_e},
\end{equation}
where $P = \{p_1, p_2, ..., p_k\}$ is a set of top-$k$ retrieved passages; $e$ is the extracted evidence; $\mathrm{L}_p$ and $\mathrm{L}_e$ denote the length of the passage $p$ and evidence $e$, respectively.
\subsection{Implementation Details} 
We employ Qwen2.5-1.5B-Instruct and Qwen2.5-7B-Instruct~\cite{qwen2025qwen25technicalreport} as the initial models. 
We optimize these two models to learn to extract rational evidence through full parameter fine-tuning for 1 epoch on the NQ dataset. 
The learning rate is $1e^{-6}$, and we select the best-performing model on the dev set. For modeling verifiable rewards, we tune $\tau$ and $\gamma$ within the ranges of $\{0.1, 0.2, 0.5, 1.0\}$ and $\{0.1, 0.3, 0.5, 0.8, 1.0\}$, respectively. 
For the length threshold $\omega$ and the weighting coefficients (\textit{i.e.,} $\lambda_1$, $\lambda_2$, and $\lambda_3$), we empirically set $\omega$, $\lambda_1$, $\lambda_2$, and $\lambda_3$ as 0.9, 0.8, 0.1, and 0.1, respectively, which commonly leads to satisfactory performance. 
For policy optimization, we set $\epsilon_{std}$ as 0.1 to stabilize training. During training, we set the PPO clipping hyperparameter $\epsilon$
and the coefficient $\beta$ controlling Kullback–Leibler (KL)-divergence as 0.2 and $1e^{-2}$, respectively.
For QA generators, we employ LoRA~\cite{hu2022lora} to fine-tune Qwen2.5-1.5B-Instruct and Qwen2.5-7B-Instruct to predict the golden answer $a$ based on the query and retrieved passages ($q$, $P$) or only the query (used for the baseline `Zero'), where LoRA rank is set as 8.
Furthermore, each experiment is run five times, and the average results are reported.
\begin{table}[ht]
  \centering
  \footnotesize
  \tabcolsep=0.14cm
  \renewcommand\arraystretch{1.05}
  \begin{tabular}{l|ccc|ccc}
    \toprule
    \multicolumn{1}{l|}{\multirow{2}{*}{\textbf{Dataset}}} & 
    \multicolumn{3}{c|}{\textbf{Recall}}  & \multicolumn{3}{c}{\textbf{NDCG}} \\
    \cline{2-7}
    & \multicolumn{1}{c}{ \multirow{1}{*}[-0.275ex]{\centering Train}} &\multicolumn{1}{c}{ \multirow{1}{*}[-0.275ex]{\centering Dev}} & \multicolumn{1}{c|}{ \multirow{1}{*}[-0.275ex]{\centering Test}} & \multicolumn{1}{c}{ \multirow{1}{*}[-0.275ex]{\centering Train}} & \multicolumn{1}{c}{ \multirow{1}{*}[-0.275ex]{\centering Dev}} & \multicolumn{1}{c}{ \multirow{1}{*}[-0.275ex]{\centering Test}} \\
    \hline
    NQ   & 78.74 & 73.07 & 74.07 & 68.30 & 61.95 & 63.08\\
    TQA & 82.35 & 77.97 & 77.77 & 76.06 & 70.32 & 70.35\\
    HotoptQA & 34.02 & 28.45 & - & 27.36 & 22.39 & -\\
    \rowcolor[HTML]{D3D3D3} 
    2WikiQA & - & - &  - & - & - & - \\
    \rowcolor[HTML]{D3D3D3} 
    Musique & - & - &  - & - & - & - \\
    \bottomrule
  \end{tabular}
  \caption{{Recall and NDCG of the top-5 retrieval passages in terms of training, development, and\, test\, sets.}}
  \label{tab:retrieval}
\end{table}
\subsection{Passage Retrieval} Following previous works \cite{DBLP:journals/corr/abs-2311-08377,DBLP:conf/acl/0025FDGY0CCC024,DBLP:conf/emnlp/ZhaoLZHCHZ24,DBLP:journals/corr/abs-2502-11811}, we retrieve Wikipedia (Wiki) passages for all datasets. We use Dense Passage Retriever
(DPR) \cite{DBLP:conf/emnlp/KarpukhinOMLWEC20} to retrieve the top-5 passages from all Wiki passages, where we use the December 2018 Wiki dump \cite{DBLP:conf/emnlp/KarpukhinOMLWEC20} as the retrieval corpus, and set the chunk size as 100 words. We use the retrieval toolkit, Tevatron \cite{DBLP:conf/sigir/GaoMLC23}, to perform corpus encoding, query encoding, and passage retrieval. Table \ref{tab:retrieval} shows the retrieval performance \textit{w.r.t.} Recall@5 and NDCG@5, where the retrieval passage that contains the golden answer is treated as the positive passage.
Besides, for HotpotQA, we compute Recall and NDCG only for the ``bridge'' questions, while ignoring yes/no and comparison questions, following previous works \cite{DBLP:journals/corr/abs-2406-15319,DBLP:conf/acl/KhalifaLLL023}.
The retrieval performance for HotpotQA is relatively low because it is a multi-hop QA dataset, and the answers are usually not spans within retrieval passages. 
Additionally, for 2WikiQA and MuSiQue, we adopt the \emph{Distractor} setting, where each question is typically equipped with two gold paragraphs and 8-9 distractor paragraphs. 
As passage retrieval is not required under this setting, we do not report retrieval performance for these two datasets.
\subsection{Comparison Baselines} To verify the effectiveness of \ours{}, we compare it with the following three groups of competitive baselines: 
\textbf{(1) Without Refinement (WR)} includes \textbf{(\rmnum{1})} Zero-shot (Zero) generates output relying on LLMs' parametric knowledge; \textbf{(\rmnum{2})} Full Passage (Full) feeds all retrieval passages into LLMs; 
\textbf{(2) Vanilla Refinement (VAR)} directly extracts evidence without explicit thinking, which includes \textbf{(\rmnum{1})} Select Context (SeleCtx) \cite{DBLP:conf/emnlp/0001DGL23} identifies and prunes redundancy in the input context based on perplexity; \textbf{(\rmnum{2})} LLMLingua-2 \cite{DBLP:conf/acl/PanWJXLZLR0LZQ024} distills compression knowledge from GPT-4 to reduce crucial information losing; \textbf{(\rmnum{3})} \textsc{Recomp}~\cite{DBLP:conf/iclr/XuSC24} selects useful sentences from retrieved documents; \textbf{(\rmnum{4})} \textsc{FilCo}~\cite{DBLP:journals/corr/abs-2311-08377} trains a context filtering model to identify key clues; \textbf{(\rmnum{5})} SEER \cite{DBLP:conf/emnlp/ZhaoLZHCHZ24} learns to extract desired evidence via self-aligned learning;
\textbf{(3) Rational Refinement (RAR)} includes \textbf{(\rmnum{1})} Chain-of-Thought (CoT) \cite{DBLP:conf/nips/Wei0SBIXCLZ22} generates query-related information from retrieval passages with explicit thinking. 
\begin{figure}[t]
    \centering  
     \subfigure[$y = \frac{1}{1+e^{-x/\tau}}$.]{
        \includegraphics[width=.475\linewidth]{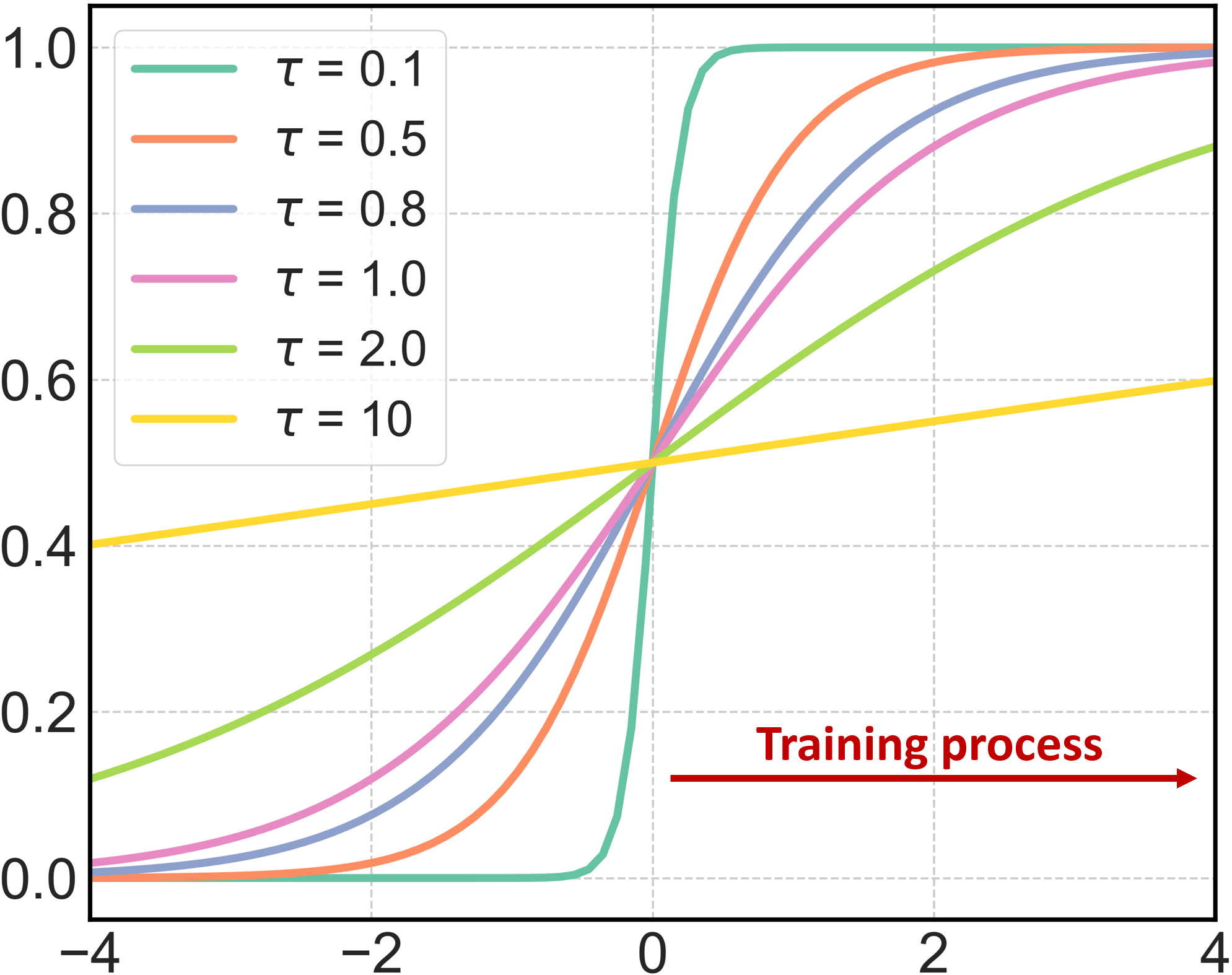}\label{fig:len_reward_r}}
    \subfigure[$y = x^{\gamma}$.]{
        \includegraphics[width=.475\linewidth]{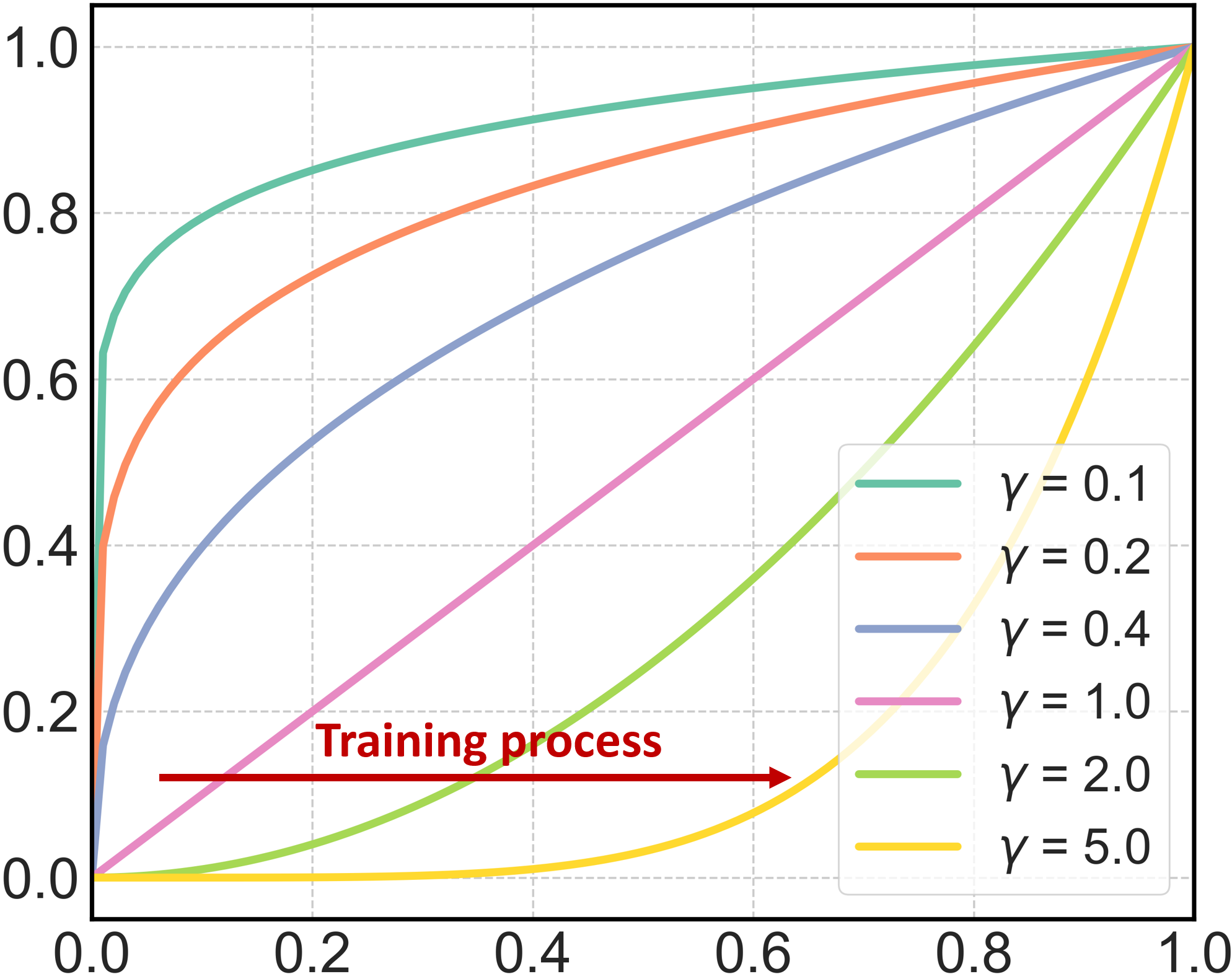}\label{fig:len_reward_e}}
    \caption{Illustration of length rewards in terms of different skewness or smoothness controlled by $\tau$ and $\gamma$, where we directly plot the basic functions for clarity.}
    \label{fig:len_reward}
\end{figure}
\section{Length Reward Formulations}
\label{app:length_def}
We define length rewards for both the rationale $r$ and the rational evidence $e$ to encourage the rationale to be comprehensive while keeping the rational evidence concise.
\smallsection{Length Reward of Rationale}  
The rationale $r$ should be sufficient to identify all key clues in the retrieved passages, serving as guidance for rational evidence extraction, which can be formulated as:
\begin{equation}
\label{equ:len_r}
    R^{len}_{r} = 
    \begin{cases}
       \sigma(\frac{\mathrm{L}_r}{\mathrm{L}_e}-1)/\tau),  & \frac{\mathrm{L}_r}{\mathrm{L}_e} \geq 1, \\
       \sigma(1-\frac{\mathrm{L}_e}{\mathrm{L}_r})/\tau), & \text{otherwise}.
    \end{cases}
\end{equation}
where $\sigma(\cdot)$ denotes the sigmoid function; $R^{len}_{r} \in (0,1]$; $\mathrm{L}_r$ and $\mathrm{L}_e$ represent the length of rationale and rational evidence, respectively.
\smallsection{Length Reward of Rational Evidence} The rational evidence $e$ should be concise relative to the retrieved passages $P$, while preserving key clues:
\begin{equation}
\label{equ:len_e}
    R^{len}_{e} = 
    \begin{cases}
        1.0, & (1-\frac{\mathrm{L}_e}{\mathrm{L}_P})\geq \omega, \\
        (1-\frac{\mathrm{L}_e}{\mathrm{L}_P})^\gamma, & \text{otherwise}.
    \end{cases}
\end{equation}
where $R^{len}_{e} \in (0,1]$; $\mathrm{L}_P$ is the length of retrieved passages;
$\tau$ and $\gamma$ are used to control the skewness or smoothness of length rewards, thereby avoiding excessively strict preferences for longer rationale or shorter rational evidence; 
$\omega$ is a threshold to avoid falling into a trivial solution.
\begin{figure*}[t]
    \centering  
     \subfigure[The workflow of agentic RAG.]{
        \includegraphics[width=.46\linewidth]{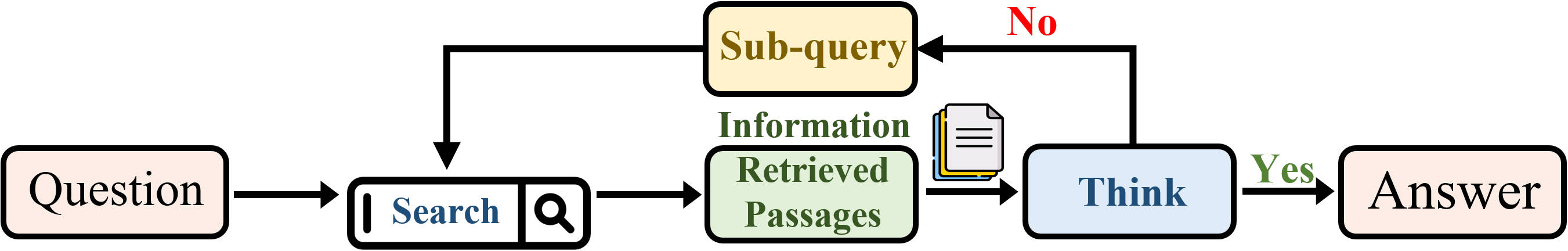}\label{fig:agen_rag}}
    \subfigure[The workflow of agentic RAG with \ours{}.]{
        \includegraphics[width=.52\linewidth]{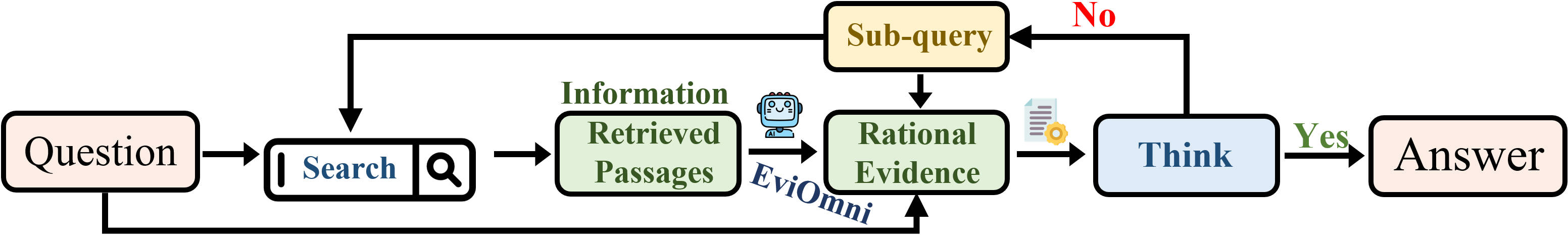}\label{fig:agen_rag_w_evi}}
    \caption{The workflow of agentic RAG (Left) and that with \ours{} (Right), where \textbf{\textcolor[RGB]{0,100,0}{``Yes''}} and \textbf{\textcolor[RGB]{200,0,0}{``No''}} indicate whether the model determines that the accumulated retrieved information or rational evidence is sufficient to answer the query, or insufficient and thus requires further search, respectively.}
    \label{fig:work_flow_agentic}
\end{figure*}
\section{Length Reward Analysis}
\label{app:length_ana}
Figure~\ref{fig:len_reward} illustrates the length rewards in terms of different skewness or smoothness controlled by $\tau$ and $\gamma$, respectively.
From Figure \ref{fig:len_reward_r}, we can see that the relatively long rationale ($x=2$) will be assigned large rewards, whereas a shorter rationale ($x=-2$) corresponds to small rewards.
On the contrary, as shown in Figure \ref{fig:len_reward_e}, shorter rational evidence ($x=0.8$) will be assigned large rewards.
By adjusting $\tau$ and $\gamma$, we can control the skewness or smoothness of length rewards, as shown in the line charts of different colors in Figure \ref{fig:len_reward}.
\begin{table}[t]
  \centering
  \footnotesize
  \tabcolsep=0.275cm
  \renewcommand\arraystretch{1.05}
  \begin{tabular}{ll|ccc}
    \toprule
    \multicolumn{2}{c|}{\textbf{Dataset}} &  \multicolumn{3}{c}{\textbf{2WikiQA}$^{\dag}$}\\
    \hline
    \multicolumn{2}{c|}{\textbf{Metric}} & EM & F1 & CR \\
    \hline
     \multicolumn{1}{l}{\multirow{2}{*}{\textbf{WR}}} & Zero & 18.69 & 49.99 & - \\
      & Full & 32.39 & 60.68 & 1.0× \\
      \hline
       \multicolumn{1}{l}{\multirow{2}{*}{\textbf{VAR}}} & LLMLingua-2 & 25.49 & 	54.77 & 4.50× \\
      & FilCo & 	27.50 & 56.58 & \underline{28.32×} \\
      \hline
       \multicolumn{1}{l}{\multirow{2}{*}{\textbf{RAR}}} & CoT & \underline{37.20} & \underline{64.44} & 	6.03× \\
      & \ours{} & \textbf{38.76} & 	\textbf{66.48} & 	\textbf{41.89×} \\
    \bottomrule
  \end{tabular}
  \caption{{Experimental Results on 2WikiQA, where ${\dag}$ denotes OOD evaluation for the proposed \ours{}.}}
  \label{tab:2wikiqa}
\end{table}
\begin{table}[t]
  \centering
  \footnotesize
  \tabcolsep=0.275cm
  \renewcommand\arraystretch{1.05}
  \begin{tabular}{ll|ccc}
    \toprule
    \multicolumn{2}{c|}{\textbf{Dataset}} &  \multicolumn{3}{c}{\textbf{MuSiQue}$^{\dag}$}\\
    \hline
    \multicolumn{2}{c|}{\textbf{Metric}} & EM & F1 & CR \\
    \hline
     \multicolumn{1}{l}{\multirow{2}{*}{\textbf{WR}}} & Zero & 1.82 & 41.92 & - \\
      & Full & 11.75  & 48.89 & 1.0× \\
      \hline
       \multicolumn{1}{l}{\multirow{2}{*}{\textbf{VAR}}} & LLMLingua-2 & 7.70 & 45.46 & 4.55× \\
      & FilCo & 8.36 & 45.74 & \underline{67.50×} \\
      \hline
       \multicolumn{1}{l}{\multirow{2}{*}{\textbf{RAR}}} & CoT & \underline{16.01} & 	\underline{53.05} & 	11.25× \\
      & \ours{} & \textbf{18.00} & 	\textbf{55.50} & \textbf{96.60×} \\
    \bottomrule
  \end{tabular}
  \caption{{Experimental Results on MuSiQue, where ${\dag}$ denotes OOD evaluation for the proposed \ours{}.}}
  \label{tab:musique}
\end{table}
\section{Additional Experiments on Multi-hop QA Benchmarks (RQ6)}
\label{app:multi_qa_bench}
To further verify the effectiveness of \ours{}, we conducted additional experiments on more challenging multi-hop QA benchmarks, \textit{i.e.,} 2WikiQA~\cite{DBLP:conf/coling/HoNSA20} and MuSiQue~\cite{DBLP:journals/tacl/TrivediBKS22}, using the Qwen2.5-1.5B-Instruct.
All experiments are conducted under the \emph{Distractor} setting, which introduces extra irrelevant documents into the context, thereby posing more challenging evaluation scenarios.
From the results shown in Table~\ref{tab:2wikiqa} and Table~\ref{tab:musique}, \ours{} achieves the best EM/F1 scores and extremely high compression ratio on both 2WikiQA and MuSiQue.
In particular, compared with `Full' and VAR methods, RAR methods consistently yield significantly better performance across multi-hop QA benchmarks, indicating that rational refinement is more effective at filtering redundant or noisy information while preserving key evidence necessary for multi-hop QA.
On the other hand, compared with CoT, \ours{} incentivizes the rational evidence extraction capability into the model itself, rather than relying on coarse chains of thought, which leads to consistently higher EM and F1 scores together with an extremely high compression ratio.
Notably, these gains are achieved under OOD evaluation settings, which further demonstrates the strong generalization of \ours{}.
These results highlight the clear advantage of \ours{} in handling complex multi-hop QA scenarios, demonstrating its superior capacity to balance generation accuracy and inference efficiency.
\begin{table*}[t]
  \centering
  \footnotesize
  \tabcolsep=0.135cm
  \renewcommand\arraystretch{1.2}
  \begin{tabular}{l|cccc|cccccc|cc}
    \toprule
    \multicolumn{1}{c|}{ \multirow{2}{*}{\textbf{Dataset}}} &  \multicolumn{4}{c|}{ \multirow{1}{*}{\textbf{Open-domain QA}}} &  \multicolumn{6}{c}{ \multirow{1}{*}{\textbf{Multi-hop QA}}} &  \multicolumn{2}{|c}{ \multirow{2}{*}{\textbf{Average}}} \\
    \cline{2-11}
     &  \multicolumn{2}{c}{\textbf{NQ}} &  \multicolumn{2}{c|}{\textbf{TQA}} &  \multicolumn{2}{c}{\textbf{HotpotQA}} &  \multicolumn{2}{c}{\textbf{2WikiQA}} &  \multicolumn{2}{c|}{\textbf{MuSiQue}} &   \\
    \hline
    \multicolumn{1}{c|}{\textbf{Metric}} & EM & CR & EM & CR & EM & CR & EM & CR & EM & CR & EM & CR\\
    \hline
    Search-R1-7B & 37.20 & 1.0x & 60.40 & 1.0x & \textbf{36.60} & 1.0x & 30.80 & 1.0x & \textbf{16.40} & 1.0x & 36.28 & 1.0x \\
    Search-R1-7B w/ \ours{} & \textbf{38.20} & \textbf{7.88x} & \textbf{63.20} & \textbf{8.29x} &  35.00 & \textbf{9.07x} & \textbf{34.80} & \textbf{10.47x} & 15.60 & \textbf{9.69x} & \textbf{37.36} & \textbf{9.08x} \\
    \hline
    Search-R1-32B & \textbf{46.60} & 1.0x & 63.80 & 1.0x & 37.20 & 1.0x & \textbf{36.60} & 1.0x & 14.40 & 1.0x & 39.72 & 1.0x \\
    Search-R1-32B w/ \ours{} & 45.40 & \textbf{8.72x} & \textbf{65.60} & \textbf{8.73x} & \textbf{37.80} & \textbf{9.29x} & 35.80 & \textbf{11.76x} & \textbf{14.80} & \textbf{9.59x} & \textbf{39.88} & \textbf{9.62x} \\
    \bottomrule
  \end{tabular}
  \caption{Performance comparison between Search-R1~\cite{DBLP:journals/corr/abs-2503-09516} with and without the proposed \ours{}.}
  \label{tab:agentic_rag_w_eviomni}
\end{table*}
\begin{table}[t]
  \centering
  \footnotesize
  \tabcolsep=0.075cm
  \renewcommand\arraystretch{1.1}
  \begin{tabular}{l|ccccc}
    \toprule
    \multicolumn{1}{c|}{\textbf{Metric}} &  \multicolumn{5}{c}{\textbf{Average number of searches}$\downarrow$} \\
    \hline
    \multicolumn{1}{c|}{\textbf{Dataset}} & NQ & TQA & HoppotQA & 2WikiQA & MuSiQue \\
    \hline
    \rowcolor[HTML]{D3D3D3} 
   \multicolumn{6}{c}{\textbf{\text{Search-R1-7B}}} \\
   \hline
     w/o \ours{} & \textbf{1.16}  & \textbf{1.20} & \textbf{1.63} & 2.19 &  2.17\\
     w/ \ours{} & 1.27 & 1.22 & 1.67 & \textbf{2.12} & \textbf{1.98} \\
     \hline
    \rowcolor[HTML]{D3D3D3} 
   \multicolumn{6}{c}{\textbf{\text{Search-R1-32B}}} \\
   \hline
     w/o \ours{} & 1.16 & 1.21 & 1.73 & 2.30 & 2.22\\
     w/ \ours{} & \textbf{1.15} & \textbf{1.15} & \textbf{1.66} & \textbf{2.06} & \textbf{2.02}\\
    \bottomrule
  \end{tabular}
  \caption{Study of the average number of searches, where lower values indicate better inference efficiency.}
  \label{tab:avg_search_cnt}
\end{table}
\section{Potential of EviOmni on Agentic RAG (RQ7)}
\label{app:poten_evi_agent}
Driven by reinforcement learning, agentic RAG models~\cite{DBLP:journals/corr/abs-2503-09516,DBLP:conf/emnlp/LiDJZZZZD25,DBLP:journals/corr/abs-2503-05592,DBLP:journals/corr/abs-2505-04588} have become an effective approach for complex QA tasks. They follow a \textcolor{blue}{think} $\rightarrow$ \textcolor{cyan}{search} $\rightarrow$ \textcolor{brown}{information} $\rightarrow$ \textcolor{blue}{think} $\rightarrow$ \textcolor{cyan}{search} $\rightarrow$ \textcolor{brown}{information} $\rightarrow$ ... $\rightarrow$ \textcolor{purple}{answer} loop, and as retrieval depth increases, accumulated context can grow rapidly, highlighting the need for a lightweight and effective denoising model to ensure both inference efficiency and generation accuracy.
To this end, we incorporate \ours{} into the agentic RAG loop as a lightweight rational evidence extractor. 
The general workflow of agentic RAG is illustrated in Figure~\ref{fig:agen_rag}, while the workflow of agentic RAG with \ours{} integration is shown in Figure~\ref{fig:agen_rag_w_evi}. 
As shown in Figure~\ref{fig:agen_rag_w_evi}, \ours{} plays a crucial role in the agentic RAG loop.
Specifically, its input query of \ours{} comprises (i) the original question, (ii) the reasoning information enclosed within \think{...}, and (iii) the sub-query enclosed within \search{...}. 
We adopt this query augmentation strategy since the sub-query alone is usually ambiguous to guide precise evidence extraction. 
At each step, the retrieved passages, together with the augmented query, are fed into \ours{} to extract rational evidence, thereby improving both the efficiency and accuracy of agentic RAG.
To evaluate the effectiveness of \ours{} in agentic RAG, we conduct experiments on a representative agentic RAG framework, \textit{i.e.,} Search-R1~\cite{DBLP:journals/corr/abs-2503-09516}. 
We consider both the 7B\footnote{\url{https://huggingface.co/PeterJinGo/SearchR1-nq_hotpotqa_train-qwen2.5-7b-it-em-grpo-v0.3}} and 32B\footnote{\url{https://huggingface.co/PeterJinGo/SearchR1-nq_hotpotqa_train-qwen2.5-32b-em-grpo-v0.3}} variants of Search-R1 to verify the robustness of \ours{} across different model scales.
For retrieval, we follow the original Searcher-R1 settings, using the 2018 Wikipedia dump~\cite{DBLP:conf/emnlp/KarpukhinOMLWEC20} as the knowledge source and E5~\cite{DBLP:journals/corr/abs-2212-03533} as the dense retriever. The maximum number of actions is set to 4.
For \ours{}, we adopt the 7B variant to serve as a lightweight rational evidence extractor within the agentic RAG loop.
We evaluate performance using Exact Match (EM) for QA accuracy and Compression Ratio (CR) to quantify the reduction of retrieved contents after applying \ours{}.
For evaluation, we randomly sample 500 instances from the test or development set of NQ, TQA, HotpotQA, 2WikiQA, and MuSiQue, respectively.
The experimental results are summarized in Table~\ref{tab:agentic_rag_w_eviomni}. We compare Search-R1 against its variant augmented with \ours{}.
The results show that integrating \ours{} consistently improves QA accuracy across both the 7B and 32B settings, yielding performance gains in 6/10 cases and comparable results in 4/10 cases, demonstrating strong generalization across both open-domain and multi-hop QA tasks.
Overall, \ours{} improves the average QA performance across five QA tasks. For example, under the 7B setting, the average EM increases from 36.28 to 37.36.
On the other hand, across both types of tasks, the average CR is approximately 9 times, substantially improving agentic RAG inference efficiency.
Compared with 32B or larger agentic RAG models, the overhead of \ours{} is negligible, making it a lightweight and practical solution for large-scale, industrial agentic RAG systems.
Additionally, we observe that \ours{} achieves higher compression ratios for multi-hop QA than for open-domain QA.
This is mainly because multi-hop QA tasks involve more search turns and thus more irrelevant retrieved contents to compress.
In conclusion, these results demonstrate that \ours{} plays a crucial role in agentic RAG systems of various scales, improving both the inference efficiency and QA accuracy across open-domain and multi-hop QA tasks.
\section{Study on Average Number of Searches (RQ7)}
\label{app:study_on_avg_search}
Table~\ref{tab:avg_search_cnt} presents the average number of searches under different settings. 
Overall, integrating \ours{} results in comparable or fewer search turns, indicating that it typically alleviates search overhead.
For the 7B setting, \ours{} leads to comparable search counts on open-domain QA datasets (\textit{i.e.,} NQ and TQA), while considerably reducing \#Avg.Searches on multi-hop QA datasets, particularly on 2WikiQA and MuSiQue.
In contrast, in the 32B setting, \ours{} consistently reduces the average number of search turns across all datasets, demonstrating more stable efficiency improvements at larger model scales.
Comparing open-domain and multi-hop QA, multi-hop tasks inherently require more search turns due to their compositional reasoning nature.
Consequently, \ours{} is more effective on multi-hop QA, where longer search chains leave greater room for eliminating redundant searches.

\section{Prompts}
\label{app:prompt}
We provide the prompts used for retrieval-augmented QA, closed-book QA, CoT generation, and rational evidence extraction in Table~\ref{tab:qa_prompt}, Table~\ref{tab:clo_prompt}, Table~\ref{tab:cot_prompt}, and Table~\ref{tab:rat_prompt}, respectively.
\section{Case Studies on Traditional RAG}
\label{app:case}
Through extensive case studies, we found the reasons for the failure of vanilla evidence mainly lie in threefold, \textit{i.e.,} \textbf{I}ncompleteness, \textbf{I}rrelevance, and \textbf{I}naccuracy, termed as ``\textbf{3I}''. 

\begin{figure*}[t]
    \centering
    \includegraphics[width=1.\linewidth]{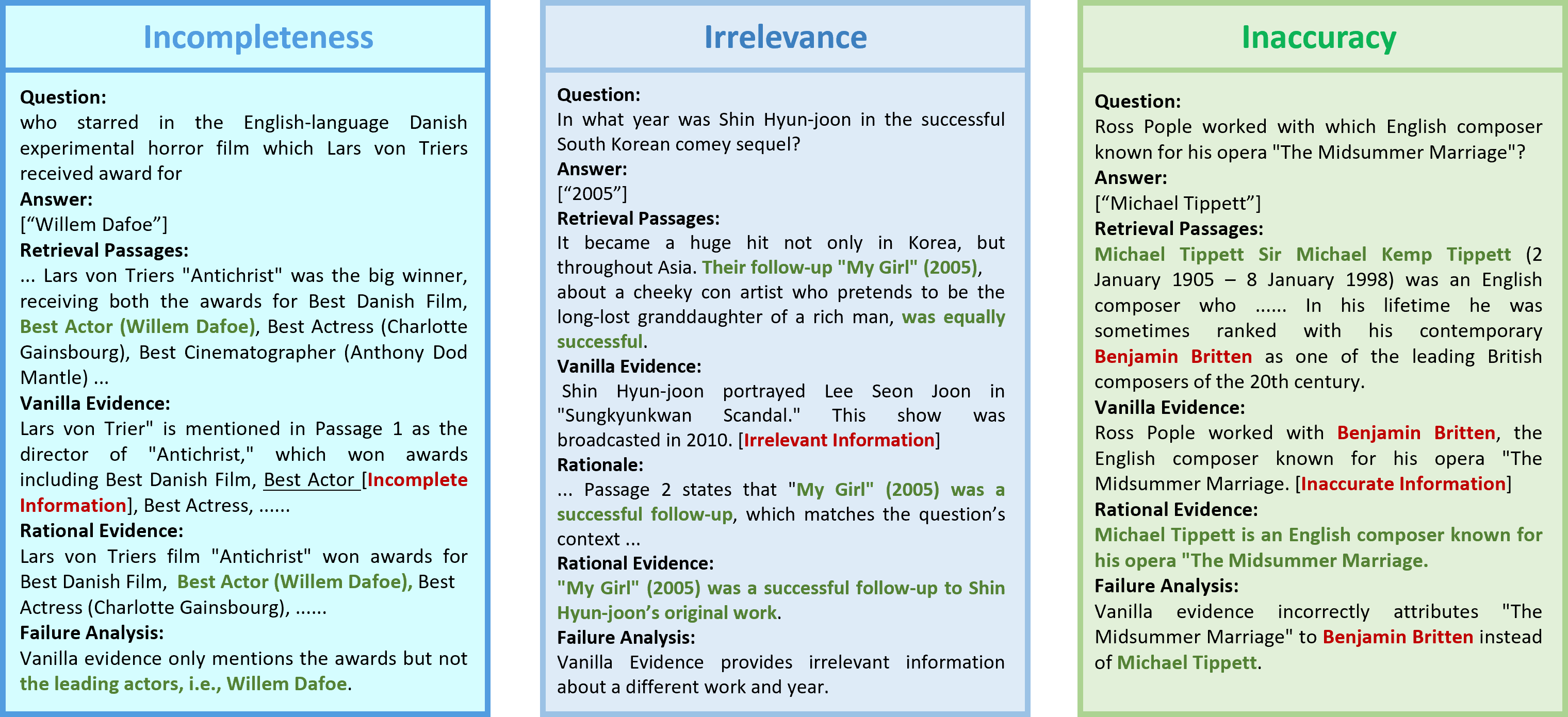}
    \caption{Three main failure issues, termed as \textbf{3I}: (1) \textbf{Incompleteness}, where the evidence provides some relevant information, but lacks the key clues needed to answer the questions; (2) \textbf{Irrelevance}, where the evidence provides information about the wrong entity, event, or topic; and (3) \textbf{Inaccuracy}, where the evidence contains incorrect information, such as wrong dates, names, or relationships.} 
    \label{fig:case_study}
\end{figure*}

\textbf{(1) Incompleteness}, where the evidence provides some relevant information but lacks key clues, an example on the left of Figure \ref{fig:case_study}, vanilla evidence provides information about the award category (Best Actor) but fails to provide the specific name of the award winner.

\textbf{(2) Irrelevance}, where the evidence provides information about the wrong entity, event, or topic, an example in the middle of Figure \ref{fig:case_study}, vanilla evidence provides irrelevant information about a different work and year.

\textbf{(3) Inaccuracy}, where the evidence contains incorrect information, such as wrong dates and names, an example in the right of Figure \ref{fig:case_study}, vanilla evidence contains a factual error, attributing the opera ``The Midsummer Marriage'' to the wrong composer (Benjamin Britten instead of Michael Tippett).

In contrast, rational evidence shows a deep contextual understanding of the query's intent, enables verification of factual accuracy, and supports reasoning to derive the correct and relevant evidence. 
It moves beyond vanilla evidence extraction towards rational evidence extraction.
\section{Case Studies on Agentic RAG}
\label{app:agen_case}
To verify the effectiveness of agentic RAG when integrated with \ours{}, we incorporate \ours{} into the agentic RAG loop as a lightweight rational evidence extractor.
We use the representative agentic RAG model Search-R1~\cite{DBLP:journals/corr/abs-2503-09516} and adopt the 7B variant of \ours{} for rational evidence extraction.
The retrieved passages are enclosed within \info{}, while rational evidence is enclosed within \infoEvi{...}.
Four cases are drawn from Search-R1, including two cases where Search-R1 succeeds and two cases where it fails.
These cases are used to analyze how \ours{} improves agentic RAG:

\smallsection{Case I (Table~\ref{tab:case1})} Integrating \ours{} ensures that only essential evidence is fed into Search-R1 at each step, substantially reducing context length and improving efficiency. Moreover, \ours{} triggers early stopping when sufficient evidence has been gathered in the second step, whereas Search-R1 continues with redundant searches. This shows that integrating \ours{} not only facilitates agentic RAG but also enables early stopping.
\smallsection{Case II (Table~\ref{tab:case2})} \ours{} precisely extracts key evidence at each step, reducing context length and improving overall efficiency. Using the rational evidence, Search-R1 generates the correct answer, demonstrating the effectiveness of rational evidence in guiding agentic RAG.
\smallsection{Case III (Table~\ref{tab:case3})} Search-R1 fails in this case, being misled by irrelevant retrieved information and producing an incorrect answer. In contrast, \ours{} explicitly detects that ``\emph{The passage does not contain any information related to The Rap Game or the contestants mentioned in the question}'' during the second step. This demonstrates that \ours{} both compresses context to improve efficiency and prevents agentic RAG from being misled by irrelevant documents.
\smallsection{Case IV (Table~\ref{tab:case4})} Search-R1 fails to generate appropriate sub-queries to decompose this complex problem, attempting to answer without sufficient evidence. In contrast, \ours{} consistently identifies that the retrieved passages lack the necessary information to answer the question. This allows the model to refuse to answer rather than produce an incorrect answer based on insufficient evidence.
In conclusion, rational evidence benefits agentic RAG through the  ``\textbf{4R}'' properties: 
\textbf{(1) Reduction} of redundant context, feeding only essential evidence at each step; 
\textbf{(2) Refusal} to answer when the available evidence is insufficient; 
\textbf{(3) Robustness} to noisy or irrelevant contents, guiding the model away from misleading information; 
and \textbf{(4) Right timing}, triggering early stopping once sufficient evidence is collected. 
When agentic RAG models succeed, integrating \ours{} leverages these properties to shorten context, accelerate inference, and avoid unnecessary search.
Conversely, when they fail, \ours{} helps identify missing/irrelevant information, preventing the model from being misled and reducing the risk of producing wrong answers.
\begin{table*}[ht]
    \renewcommand{\arraystretch}{1.25} 
    \begin{tcolorbox}[
        sharp corners,
        colback=gray!5,
        colframe=black!80,
        fonttitle=\bfseries,
        title=Prompt for Retrieval-Augmented QA,
        boxrule=0.99pt,
        arc=0mm
    ]
            \textcolor{black}{\textbf{[Instruction]}}\\
            You are a helpful assistant. Your task is:
            \begin{enumerate}[itemsep=1pt, parsep=0pt, topsep=3pt, partopsep=0pt, leftmargin=1.5em]
                \item Read the given question and use the documents provided to answer the question.
                \item If the documents don't work, please answer the question based on your own knowledge.
            \end{enumerate}
            {Question:} {\{question\}} \\
            {Document:} {\{document\}} \\
            {Answer:} 
    \end{tcolorbox}
    \caption{The prompt for retrieval-augmented QA.}
    \label{tab:qa_prompt}
\end{table*}
\begin{table*}[ht]
    \renewcommand{\arraystretch}{1.25} 
    \begin{tcolorbox}[
        sharp corners,
        colback=gray!5,
        colframe=black!80,
        fonttitle=\bfseries,
        title=Prompt for Closed-book QA,
        boxrule=0.99pt,
        arc=0mm
    ]
            % \textbf{Prompt for Closed-book QA} 
            % \tcblower
            \textcolor{black}{\textbf{[Instruction]}}\\
            You are a helpful assistant. Your task is:
            \begin{enumerate}[itemsep=1pt, parsep=0pt, topsep=3pt, partopsep=0pt, leftmargin=1.5em]
                \item  Read the given question and then answer the question directly.
                \item  Give a short answer to the question based on your own knowledge.
            \end{enumerate}
            {Question:} {\{question\}} \\
            {Answer:} 
    \end{tcolorbox}
    \caption{The prompt for closed-book QA.}
    \label{tab:clo_prompt}
\end{table*}
\begin{table*}[ht]
    \renewcommand{\arraystretch}{1.25} 
    \begin{tcolorbox}[
        sharp corners,
        colback=gray!5,
        colframe=black!80,
        fonttitle=\bfseries,
        title=Prompt for CoT Generation,
        boxrule=0.99pt,
        arc=0mm
    ]
            % \textbf{Prompt for CoT Generation} 
            % \tcblower
            \textcolor{black}{\textbf{[Instruction]}}\\
            You are a helpful assistant. Your task is: \\
            Read the given documents, and answer the question below. \\
            {Question:} {\{question\}} \\
            {Document:} {\{document\}} \\
            Let's think step by step.
    \end{tcolorbox}
    \caption{The prompt for Chain-of-Thought generation.}
    \label{tab:cot_prompt}
\end{table*}
\begin{table*}[ht]
    \renewcommand{\arraystretch}{1.2} 
    \begin{tcolorbox}[
        sharp corners,
        colback=gray!5,
        colframe=black!80,
        fonttitle=\bfseries,
        title=Prompt for Rational Evidence Extraction,
        boxrule=0.99pt,
        arc=0mm
    ]
            \textcolor{black}{\textbf{[Instruction]}}\\
            You are a highly skilled knowledge reasoner and extractor. \\
            Your task is to carefully read the given question and passages to reason how the passages lead to the answer and extract relevant information that may be used to answer the question. \\
            \textbf{Follow these steps:} 
            \begin{enumerate}[label=\arabic*., itemsep=1pt, parsep=0pt, topsep=3pt, partopsep=0pt, leftmargin=1.5em]
                \item In the \textcolor[RGB]{68,114,196}{\textless reason\textgreater}\textcolor[RGB]{68,114,196}{\textless /reason\textgreater} tag, perform the following steps. Question Analysis: Analyze the question to understand the specific information they are seeking. Identify the key concepts, entities, and relationships involved. Passage Analysis: For each passage, carefully read and identify sentences or phrases that are useful for answering the given question.
                \item In the \textcolor[RGB]{0,100,0}{\textless extract\textgreater}\textcolor[RGB]{0,100,0}{\textless /extract\textgreater} tag, synthesize useful information from the passages into a coherent narrative. Organize the information logically and concisely.
                \item In \textcolor[RGB]{127,100,0}{\textless answer\textgreater}\textcolor[RGB]{127,100,0}{\textless /answer\textgreater} tags, give a short answer to the given question, based on the passages, reasoning information, and extracted knowledge. If none of them work, please answer the question based on your knowledge.
            \end{enumerate}
            {Question:} {\{question\}} \\
            {Passages:} {\{passages\}} \\
    \end{tcolorbox}
    \caption{The prompt for rational evidence extraction, where generation is terminated when encountering token `\textless /extract\textgreater'.}
    \label{tab:rat_prompt}
\end{table*}

\begin{table*}[h]
    \centering
    \small 
    \renewcommand{\arraystretch}{1.25}
    \begin{tabular}{p{15.5cm}}
        \hline
        \textbf{Question}: Curious is a women's fragrance by a singer born in what city and state? \\
        \hline
        \textbf{Ground Truth}: McComb, Mississippi \\
        \hline
        \textbf{Step 1}: \\
        \think{I need to find out which city and state a singer, by whom the women's fragrance "Curious" was created, was born in.} \\
        \search{Curious fragrance information} \\
        \info{Doc 1(Title: "Curious (fragrance)") Curious (fragrance) Curious is a women's fragrance by Britney Spears for Elizabeth Arden, and is the first perfume to be endorsed by Britney Spears, preceding ""Fantasy."" Spears reportedly earned \$52 million from the endorsement. ""Curious"" was released in September 2004, and was very successful internationally. It was the number one fragrance of 2004 in department stores, and in 2005, ""Curious"" was honored by the Fragrance Foundation as Best Women's Fragrance. The fragrance is a white floral scent. Its notes are Louisiana magnolia, golden Anjou pear, lotus flower, tuberose, star jasmine, pink cyclamen, vanilla-infused musk, sandalwood and blonde woods. Elizabeth Arden ...
        } \\
        \infoEvi{\textbf{Britney Spears created the women’s fragrance "Curious" for Elizabeth Arden.} However, the passage does not provide information about Britney Spears' birthplace.} \\
        \hline
        \textbf{Step 2}: \\
        \think{I need to find out which city and state Britney Spears, the singer behind "Curious" fragrance, was born in.} \\
        \search{Britney Spears birthplace} \\
        \info{Doc 1(Title: "Britney Spears") the best-selling female albums artist of the 2000s. In 2004, she launched a perfume brand with Elizabeth Arden, Inc., from which sales exceeded US\$1.5 billion, . Spears was born in McComb, Mississippi, the second child of Lynne Irene Bridges and James Parnell Spears. Her maternal grandmother, Lillian Portell, was English (born in London), and one of Spears's maternal great-great-grandfathers was Maltese. Her siblings are Bryan James and Jamie Lynn. Britney was born in the Bible Belt, where socially conservative evangelical Protestantism is a particularly strong religious influence. Spears was baptized into the Southern Baptist Convention, but later in life studied ...
        } \\
        \infoEvi{\textbf{Britney Spears was born in McComb, Mississippi.} This information is explicitly stated in the passage, providing the city and state of her birth.} \\
        \textit{\textbf{Terminate early because \ours{} believes that sufficient evidence has been obtained}}\\
        \hline
        \textbf{Step 3}: \\
        \think{Since Britney Spears was born in McComb, Mississippi, I need to find out which city and state that is.} \\
        \search{McComb, Mississippi location} \\
        \info{Doc 1(Title: "McComb, Mississippi") McComb, Mississippi McComb is a city in Pike County, Mississippi, United States, approximately south of Jackson. As of the 2010 census, the city had a total population of 12,790. It is the principal city of the McComb, Mississippi Micropolitan Statistical Area. McComb was founded in 1872 after Henry Simpson McComb of the New Orleans, Jackson and Great Northern Railroad, a predecessor of the Illinois Central Railroad (now part of the Canadian National Railway), decided to move the railroad's maintenance shops away from New Orleans, Louisiana, to avoid the attractions of that city's saloons. The railroad purchased land in Pike County, ...
        } \\
        \hline
        \think{I have found that Britney Spears was born in McComb, Mississippi. Since the question asks for the city and state, I just need to confirm the state, which I already know from the previous search.} \\
        \answer{McComb, Mississippi}\\
                \hline
    \end{tabular}
    \caption{Agentic RAG with \ours{} case study I, where \ours{} triggers early stopping when it judges the evidence sufficient.}\label{tab:case1}
\end{table*}

\begin{table*}[h]
    \centering
    \small
    \renewcommand{\arraystretch}{1.0}
    \begin{tabular}{p{15.5cm}}
        \hline
        \textbf{Question}: Where are Teide National Park and Garajonay National Park located? \\
        \hline
        \textbf{Ground Truth}: Canary Islands, Spain \\
        \hline
        \textbf{Step 1}: \\
        \think{I need to find out where Teide National Park and Garajonay National Park are located. I'll search for it.} \\
        \search{Teide National Park and Garajonay National Park location} \\
        \info{Doc 1(Title: "Garajonay National Park") decided to take their own lives. Garajonay National Park Garajonay National Park (, ) is located in the center and north of the island of La Gomera, one of the Canary Islands (Spain). It was declared a national park in 1981 and a World Heritage Site by UNESCO in 1986. It occupies 40 km (15 sq mi) and it extends into each of the six municipalities on the island. Nearly 450,000 tourists visit the park every year. There is accommodation for tourists just outside the park's boundary. The park is named after the rock formation of Garajonay, the highest point ...
        } \\
        \infoEvi{Garajonay National Park is located in the center and north of the island of La Gomera, \textbf{one of the Canary Islands (Spain).}} \\
        \hline
        \textbf{Step 2}: \\
        \think{I found out that Garajonay National Park is located in the center and north of the island of La Gomera, one of the Canary Islands (Spain). Now I need to find the location of Teide National Park.} \\
        \search{Teide National Park location} \\
        \info{Doc 1(Title: "Teide National Park") Teide National Park Teide National Park (, ) is a national park located in Tenerife (Canary Islands, Spain). The national park is centered on Mount Teide, the highest mountain of Spain (3,718 meters high) in addition, the highest volcano in the Atlantic Ocean islands and the third highest volcano in the world from its base in the ocean floor (7.500 meters high). Its national park status was passed on 22 January 1954, making it the third oldest national park in Spain. Pico Viejo, also included within the national park limits, is the second highest volcano in the Canary Islands with ...
        } \\
        \infoEvi{Teide National Park is located in Tenerife, \textbf{one of the Canary Islands, Spain.} The park is centered on Mount Teide, the highest mountain in Spain.} \\
        \hline
        \think{I found out that Teide National Park is located in Tenerife (Canary Islands, Spain). Now I have the locations of both Teide National Park and Garajonay National Park.} \\
        \answer{Canary Islands, Spain}\\
        \hline
        \end{tabular}
\caption{Agentic RAG with \ours{} case study II, where \ours{} efficiently extracts key evidence at each step, reducing redundant context and improving generation efficiency.}\label{tab:case2}
\end{table*}

\begin{table*}[h]
    \centering
    \small
    \renewcommand{\arraystretch}{1.0}
    \begin{tabular}{p{15.5cm}}
        \hline
        \textbf{Question}: Who was the winner of the season of The Rap Game where a contestant named Lady Leshurr as one of her influences? \\
        \hline
        \textbf{Ground Truth}: Mani \\
        \hline
        \textbf{Step 1}: \\
        \think{I need to find the winner of the season of The Rap Game where a contestant named Lady Leshurr was one of her influences. I'll search for it.} \\
        \search{who was the winner of the season of The Rap Game where a contestant named Lady Leshurr as one of her influences} \\
        \info{Doc 1(Title: "Lady Leshurr") Act. Leshurr came out as pansexual in September 2018. Leshurr has named Ashley Walters, Lisa Maffia, Missy Elliott, and Lil Wayne as her primary influences. Lady Leshurr Melesha Katrina O'Garro (born 15 December 1989) known professionally as Lady Leshurr (), is a British grime and hip hop rapper, singer, and producer from Kingshurst, Solihull, UK. She is best known for her ""Queen's Speech"" series of freestyles, the fourth of which went viral in 2016. ""The Rap Game"" season 2 competitor Nia Kay stated that Leshurr was one of her influences due to her uniqueness and rap skill. Her subsequent freestyle, ...
        } \\
        \infoEvi{\textbf{Lady Leshurr was an influence on Nia Kay,} who competed in season 2 of The Rap Game. However, the winner of that season is not directly mentioned in the provided passages.}\\
        \hline
        \textbf{Step 2}: \\
        \think{I found out that Lady Leshurr was one of the influences for Nia Kay in The Rap Game season 2. Now I need to find the winner of that season.} \\
        \search{winner of The Rap Game season 2} \\
        \info{Doc 1(Title: "Chandra Davis") Chandra Davis Chandra Davis (born January 2, 1978 in Detroit, Michigan), a.k.a. London Charles and Deelishis, is an American model, and former reality show contestant. She is best known as the winner of VH1's hit reality TV show ""Flavor of Love 2"". Davis was a contestant on the second season of VH1's Flavor of Love (aged 29). She went on to win the competition when Flavor Flav chose her over Tiffany 'New York' Pollard, who returned after previously named the runner-up on the show's first season. After FOL, Davis announced she is currently not dating Flav, via a VH1 special, ...
        } \\
        \infoEvi{\textbf{The passage does not contain any information related to The Rap Game or the contestants mentioned in the question.} It only discusses Chandra Davis, who won "Flavor of Love 2.}\\
        \hline
        \think{I found out that the winner of The Rap Game season 2 is London Charles (Chandra Davis). Now I can provide the answer.} \\
        \answer{London Charles}\\
                \hline
            \end{tabular}
    \caption{Agentic RAG with \ours{} case study III, where \ours{} explicitly points out that ``The passage does not contain any information related to The Rap Game or the contestants mentioned in the question.'' in the second step.}\label{tab:case3}
\end{table*}

\begin{table*}[h]
    \centering
    \small
    \renewcommand{\arraystretch}{1.25}
    \begin{tabular}{p{15.5cm}}
        \hline
        \textbf{Question}: This expansion of the 2008 magazine article "Is Google Making Us Stoopid?" was a finalist for what award? \\
        \hline
        \textbf{Ground Truth}: Pulitzer Prize \\
        \hline
        \textbf{Step 1}: \\
        \think{I need to find out which expansion of the 2008 magazine article "Is Google Making Us Stoopid?" was a finalist for what award. I'll search for it.} \\
        \search{This expansion of the 2008 magazine article "Is Google Making Us Stoopid?" was a finalist for what award} \\
        \info{Doc 1(Title: "Is Google Making Us Stupid?") Is Google Making Us Stupid? Is Google Making Us Stupid? What the Internet Is Doing to Our Brains (alternatively Is Google Making Us Stoopid?) is a magazine article by technology writer Nicholas G. Carr, and is highly critical of the Internet's effect on cognition. It was published in the July/August 2008 edition of ""The Atlantic"" magazine as a six-page cover story. Carr's main argument is that the Internet might have detrimental effects on cognition that diminish the capacity for concentration and contemplation. Despite the title, the article is not specifically targeted at Google, but more at the cognitive impact of ...
        } \\
        \infoEvi{The article "Is Google Making Us Stoopid?" by Nicholas G. Carr was published in the July/August 2008 edition of "The Atlantic" magazine. \textbf{However, the passage does not mention any awards the article was a finalist for.}} \\
        \hline
        \textbf{Step 2}: \\
        \think{I found out that the expansion of the 2008 magazine article "Is Google Making Us Stoopid?" was a finalist for an award. I need to find out which award it was finalist for.} \\
        \search{which award was the expansion of the 2008 magazine article "Is Google Making Us Stoopid?" finalist for} \\
        \info{Doc 2(Title: "Is Google Making Us Stupid?") inconclusive and stated that ""it will take time before it's clear whether we should mourn the old ways, celebrate the new, or learn to stop worrying and love the Net"". Is Google Making Us Stupid? Is Google Making Us Stupid? What the Internet Is Doing to Our Brains (alternatively Is Google Making Us Stoopid?) is a magazine article by technology writer Nicholas G. Carr, and is highly critical of the Internet's effect on cognition. It was published in the July/August 2008 edition of ""The Atlantic"" magazine as a six-page cover story. Carr's main argument is that the Internet might have ...
        } \\
        \infoEvi{The passages \textbf{do not provide information about the expansion of the 2008 article or the award it was a finalist for.} However, it is common for such articles to be nominated for literary awards like the National Magazine Award.} \\
        \hline
        \textbf{Step 3}: \\
        \think{I found out that the expansion of the 2008 magazine article "Is Google Making Us Stoopid?" was a finalist for an award. I need to find out which award it was finalist for.} \\
        \search{which award was the expansion of the 2008 magazine article "Is Google Making Us Stoopid?" finalist for} \\
        \info{Doc 1(Title: "Is Google Making Us Stupid?") Is Google Making Us Stupid? Is Google Making Us Stupid? What the Internet Is Doing to Our Brains (alternatively Is Google Making Us Stoopid?) is a magazine article by technology writer Nicholas G. Carr, and is highly critical of the Internet's effect on cognition. It was published in the July/August 2008 edition of ""The Atlantic"" magazine as a six-page cover story. Carr's main argument is that the Internet might have detrimental effects on cognition that diminish the capacity for concentration and contemplation. Despite the title, the article is not specifically targeted at Google, but more at the cognitive impact of ...
        } \\
        \infoEvi{\textbf{The passage does not provide any information about the award the expansion of the 2008 magazine article "Is Google Making Us Stoopid?" was a finalist for.} The article discusses the original publication details but does not mention any awards.} \\
        \hline
        \answer{National Magazine Award}\\
        \hline
    \end{tabular}
    \caption{Agentic RAG with \ours{} case study IV, where \ours{} consistently highlights in its extracted rational evidence that the passages do not provide enough information to answer the question.}\label{tab:case4}
\end{table*}

\clearpage

\renewcommand{\algorithmicrequire}{\textbf{Input:}}  
\renewcommand{\algorithmicensure}{\textbf{Output:}}  
\begin{algorithm*}[t]
\caption{Traditional RAG with \ours{}}
\label{alg:rag_eviomni}
\begin{algorithmic}[1]
\Require Input query $q$, retriever $\mathcal{R}$, generator $\mathcal{M}_\mathcal{G}$, \ours{} model $\mathcal{M}_\mathcal{E}$
\Ensure Final answer $o$
    \State Retrieve relevant passages $P \gets \mathcal{R}(q)$
    \State Generate rationale $r \gets \mathcal{M}_\mathcal{E}(\cdot|q, P)$ 
    \State Generate rational evidence $e \gets \mathcal{M}_\mathcal{E}(\cdot|q, P, r)$ 
    \State Generate final answer $o \gets \mathcal{M}_\mathcal{G}(\cdot |q, e)$ 
    \State \Return $o$
\end{algorithmic}
\label{alg:trad_rag}
\end{algorithm*}

\begin{algorithm*}[t]
\caption{Agentic RAG with \ours{}}
\label{alg:agentic_rag_eviomni}
\begin{algorithmic}[1]
\Require Input query $q$, max iterations $T$, retriever $\mathcal{R}$, agentic RAG model $\mathcal{M}_\mathcal{G}$, \ours{} model $\mathcal{M}_\mathcal{E}$
\Ensure Final answer $o$
    \State Initialize history buffer $y \gets \emptyset$ 
    \State Initialize action count $t \gets 0$
    \State Initialize final answer $o \gets \text{none}$
     \While{$t<T$}
        \State Conduct step-by-step reasoning $y_t \gets \mathcal{M}_\mathcal{G}(\cdot|q, y)$ 
        \State Append $y_t$ to the history buffer $y \gets y+y_t$
        \State Generate next action $y_a \gets \mathcal{M}_\mathcal{G}(\cdot|q, y)$ 
        \If{<search> </search> tag in $y_a$}
            \State Extract sub-query $q_s$ from $y_a$
            \State Retrieve relevant passages $P \gets \mathcal{R}(q_s)$
            \State Compose context-aware query $q_c \gets q\oplus y_t \oplus q_s$ \Comment{Integrate the original query, reasoning information, and sub-query to form context-aware query with high contextual integrity}
            \State Generate rationale using \ours{} $r \gets \mathcal{M}_\mathcal{E}(\cdot|q_c, P)$
            \State Generate rational evidence using \ours{} $y_e \gets \mathcal{M}_\mathcal{E}(\cdot|q_c, P, r)$
            \State Append $y_a$ and $y_e$ to the history buffer $y \gets y+y_a+y_e$
        \EndIf
        \If{<answer> </answer> tag in $y_a$}
            \State Extract answer $o_t$ from $y_a$
            \State Assign $o_t$ to the final answer $o \gets o_t$
            \State break
        \EndIf
        \State Increment action count $t \gets t+1$
    \EndWhile
    \State \Return $o$
\end{algorithmic}
\label{alg:agent_rag}
\end{algorithm*}

\end{document}